\documentclass{article}

\usepackage{microtype}
\usepackage{graphicx}
\usepackage{subcaption}
\usepackage{float}
\usepackage{booktabs}
\usepackage{hyperref}
\usepackage{xurl}
\hypersetup{breaklinks=true}

\usepackage[preprint]{icml2026}

\usepackage{amsmath}
\usepackage{amssymb}
\usepackage{mathtools}
\usepackage{amsthm}
\usepackage{amsfonts}
\usepackage{bm}
\usepackage{dsfont}

\usepackage{xcolor}
\usepackage{pifont}
\usepackage{tabularx}
\usepackage{enumitem}
\usepackage[capitalize,noabbrev]{cleveref}

\def\1{\bm{1}}

\def\vp{{\bm{p}}}

\def\vr{{\bm{r}}}

\def\vw{{\bm{w}}}
\def\vx{{\bm{x}}}

\def\mI{{\bm{I}}}

\def\mW{{\bm{W}}}

\DeclareMathAlphabet{\mathsfit}{\encodingdefault}{\sfdefault}{m}{sl}
\SetMathAlphabet{\mathsfit}{bold}{\encodingdefault}{\sfdefault}{bx}{n}

\newcommand{\E}{\mathbb{E}}

\DeclareMathOperator*{\argmax}{arg\,max}
\DeclareMathOperator*{\argmin}{arg\,min}

\newcommand{\mc}{\mathcal}

\newcommand{\xmark}{\ding{55}}

\icmltitlerunning{On the Trainability of Masked Diffusion Language Models via Blockwise Locality}

\begin{document}








\twocolumn[
  \icmltitle{On the Trainability of Masked Diffusion Language Models via Blockwise Locality}

  \begin{icmlauthorlist}
    \icmlauthor{Yuxiang Wang$^*$}{fudan}
    \icmlauthor{Yu Xiang$^*$}{fudan}
    \icmlauthor{Baojian Zhou}{fudan}
    \icmlauthor{Qifang Zhao}{alibaba}
    \icmlauthor{Keyue Jiang}{alibaba}
    \icmlauthor{Yanghua Xiao}{fudan}
    \icmlauthor{Xiaoxiao Xu}{alibaba}
  \end{icmlauthorlist}

  \icmlaffiliation{fudan}{Fudan University, Shanghai, China}
  \icmlaffiliation{alibaba}{Alibaba Group, China}

  \icmlcorrespondingauthor{Yuxiang Wang}{25210980109@m.fudan.edu.cn}

  \icmlkeywords{Diffusion language models, In-context learning, Masked diffusion language models}

  \vskip 0.3in
]

\printAffiliationsAndNotice{\icmlEqualContribution}


\begin{abstract}
Masked diffusion language models (MDMs) have recently emerged as a promising alternative to standard autoregressive large language models (AR-LLMs), yet their optimization can be substantially less stable. We study blockwise MDMs and compare them with AR-LLMs on three controlled tasks that stress different aspects of structured generation: in-context linear regression, graph path-finding, and Sudoku solving. We find that standard random-masking MDMs fail to reliably learn linear regression, exhibit high-variance training dynamics on graph path-finding, while outperforming AR-LLMs on Sudoku. To mitigate these instabilities, we propose two locality-aware blockwise models, namely Jigsaw and Scatter, that inject \textit{left-to-right inductive bias} by enforcing autoregressive locality \emph{within} blocks while preserving iterative refinement at the block level. Empirically, Jigsaw matches AR-LLM stability on linear regression and remains strong on Sudoku, while Scatter retains diffusion's planning advantage on path-finding. Our results indicate that standard random-masking MDMs, even with blockwise variants, may be a suboptimal instantiation of diffusion LMs for ordered generation, motivating models beyond random masking.
\end{abstract}

\section{Introduction}
\label{sect:intro}

\begin{figure}[ht]
\centering
\includegraphics[width=.95\linewidth]{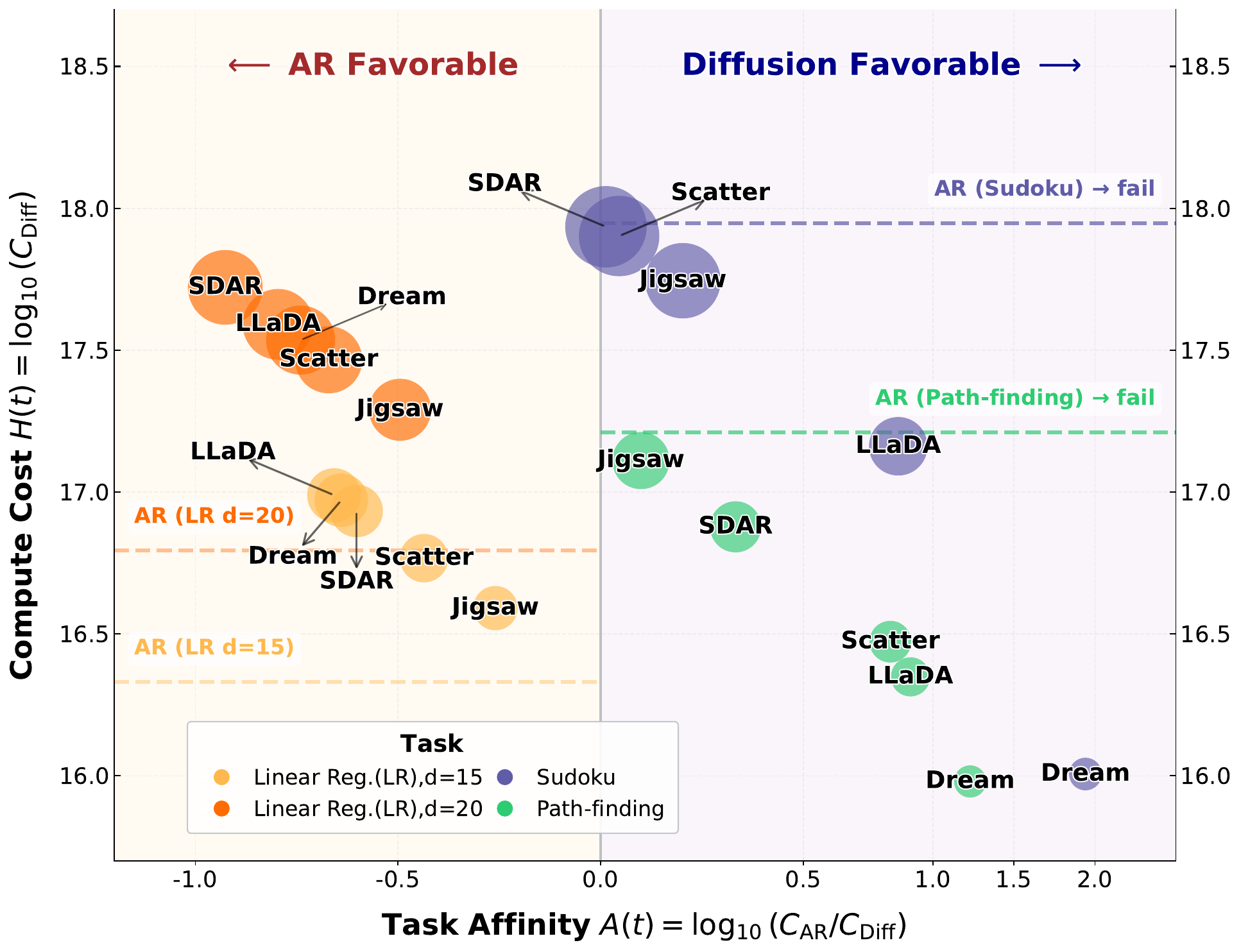}
\caption{Task affinity between AR-LLMs and MDMs. The $x$-axis reports 
$A(t)=\log_{10}\!\big(\tfrac{C_{\mathrm{AR}}(t;\tau)}{C_{\mathrm{Diff}}(t;\tau)}\big)$, 
where $C(t;\tau)$ denotes the cumulative training FLOPs required to reach a task-specific threshold $\tau$. 
$A(t)<0$ indicates AR-favorable tasks, while $A(t)>0$ indicates MDMs are more compute-efficient. 
Our proposed Jigsaw and Scatter significantly reduce the compute-to-target for linear regression, narrowing the trainability gap. 
To make the comparison transparent, $C(t;\tau)$ is estimated using profiling-based FLOPs accounting: for each task and paradigm, we measure the actual train-step FLOPs with PyTorch profiling under matched architectures, sequence lengths, and batch sizes, and then accumulate FLOPs until the corresponding threshold is reached. 
When $\tau$ falls between checkpoints, we use log-space interpolation. 
Task-specific thresholds and additional profiling details are provided in Appendix~\ref{appendx-sect:fig-details}.
\vspace{-6mm}
}
\label{fig:affinity-hardness-AR-LLMs-vs-dLLMs}
\end{figure}

Autoregressive large language models (AR-LLMs) have become the dominant paradigm in modern AI systems due to their remarkable language understanding and generation abilities. However, their planning capability remains limited, partly because inference is constrained to a fixed left-to-right factorization \cite{valmeekam2023planning,wang2024alpine}. This has motivated growing interest in alternative generation paradigms, including diffusion-based language models (dLLMs) \cite{austin2021structured,lou2024discrete,gat2024discrete,sahoo2024simple,shi2024simplified,ou2025your,arriola2025block,nie2025large,ye2025dream}, which generate via iterative denoising and thus differ fundamentally from teacher-forced autoregression.

Among dLLMs, masked diffusion models (MDMs) have been reported to achieve performance comparable to AR-LLMs on a range of benchmarks \cite{sahoo2024simple,shi2024simplified,ou2025your}. Unlike autoregressive decoding, MDMs generate by iteratively denoising randomly masked tokens, which enables parallel token updates and provides a natural pathway to iterative refinement. While dLLMs can excel on planning-style problems such as Sudoku, they often lag behind or show limited advantages on natural language tasks, including mathematical reasoning \cite{ye2025dream,nie2025large,ye2025beyond}.  We hypothesize that this mixed performance is driven by differences in \textit{inductive bias}. Many language understanding and math reasoning benchmarks strongly favor left-to-right dependency modeling, which aligns naturally with AR training and decoding, whereas MDMs are trained via denoising objectives that do not impose a fixed directional factorization \cite{du2025understanding}. In contrast, planning-style problems such as Sudoku and path-finding emphasize discovering or refining a global token configuration with weak left-to-right structure, where iterative denoising can be advantageous as shown in Fig. \ref{fig:affinity-hardness-AR-LLMs-vs-dLLMs}. This raises the following question:

\begin{center}
\textit{When does the autoregressive left-to-right inductive bias help relative to masked diffusion denoising, and can we design a model that combines the strengths of both?}
\end{center}

While recent works \cite{han2023ssd,cheng2025sdar,arriola2025block,fathi2025unifying} explore block-structured architectures that combine elements of AR-LLMs and MDMs, these hybrids still do not explicitly inject the left-to-right \textit{locality} inductive bias that is central to autoregressive modeling. Most existing designs perform diffusion \emph{within} each block while generating blocks \emph{autoregressively} across the sequence. For natural language, however, left-to-right dependencies are primarily token-local; diffusing within a block can blur token-level left-to-right dependencies, making it harder to exploit the locality bias that benefits language modeling. Motivated by this, we propose two locality-aware blockwise diffusion models, namely \textbf{Jigsaw} and \textbf{Scatter}, that impose left-to-right order at the block diffusion level while preserving token-level locality. Empirically, these locality-aware designs improve stability and compute-to-target on tasks with strong local left-to-right structure (notably linear regression), while maintaining strong performance on globally constrained generation (Sudoku), and also revealing when locality can conflict with reverse-dependency planning (path-finding). 

To summarize, 
\vspace{-2mm}
\begin{itemize}[leftmargin=*]
\item We systematically evaluate the trainability of standard MDMs and block-structured variants on three controlled tasks (in-context linear regression, star-graph path-finding, and Sudoku) that isolate ordered binding, reverse planning, and global constraint satisfaction. We find that standard random-masking MDMs are difficult to optimize on linear regression and exhibit high-variance dynamics on path-finding, while AR-LLMs fail on Sudoku, underscoring a sharp dependence on inductive bias.

\item We propose two locality-aware blockwise diffusion architectures, Jigsaw and Scatter, that inject token-local left-to-right structure by enforcing autoregressive generation within blocks while retaining diffusion-style refinement across blocks (via entropy-guided planning in Jigsaw and synchronized parallel autoregression in Scatter).

\item Experiments show that these designs improve the tradeoff between trainability and stability relative to standard MDM baselines and prior blockwise variants: they reduce variance and compute-to-target on ordered binding tasks, retain strong performance on Sudoku, and provide diagnostic evidence that aligning locality with task dependency structure is crucial\footnote{Code is available at \url{https://github.com/stein-wang0226/trainable-masked-diffusion}}.
\end{itemize}

\subsection{Related work}
\label{sect:related-work}

\textbf{Limitations of left-to-right autoregression.} The limitations of left-to-right generation have been analyzed from both optimization and inference perspectives. \citet{bachmann2024pitfalls} argue that teacher forcing can encourage ``Clever Hans'' shortcuts, where models rely on correct prefixes rather than learning robust lookahead in graph path-finding task. Complementarily, the ``Factorization Curse'' \cite{kitouni2024factorization} highlights that fixed left-to-right factorization can hinder learning flexible joint distributions, and inference-time error accumulation further exacerbates brittleness under long-horizon dependencies.

\textbf{Discrete diffusion language models.} Diffusion models for text \cite{hoogeboom2021argmax,zheng2023diffusionbert,austin2021structured,lou2024discrete} generate sequences via iterative denoising, enabling parallel token updates and global refinement. Recent masked diffusion language models (MDMs) have reported competitive results with AR-LLMs on several benchmarks \cite{sahoo2024simple,shi2024simplified,ou2025your}, and diffusion-style objectives have also been explored for improved global consistency and planning \cite{ye2025beyond,nie2025large,ye2025dream}. Despite these advances, diffusion LMs often lag behind AR models on natural-language reasoning benchmarks, suggesting a task-dependent gap that may be tied to inductive bias.

\textbf{Block-structured diffusions.} Several works explore block-structured architectures that combine diffusion and autoregressive components \cite{han2023ssd,cheng2025sdar,arriola2025block,fathi2025unifying,ma2026diffdiff}. Most existing designs diffuse \emph{within} blocks while generating blocks \emph{autoregressively} across the sequence, yielding a coarse directional structure. Our work is motivated by the observation that for language tasks, the useful left-to-right inductive bias is often token-local; thus, injecting locality at the appropriate granularity is critical for stable training.

\textbf{Related works on inference optimization and ordering in diffusion LMs.} Several works improve MDMs \emph{only at inference time} by optimizing the sampling dynamics. \citet{chen2025optimizing} propose entropy-guided search to reduce path uncertainty; \citet{luxembourg2025plan} use dilated schedules to cut the number of steps to logarithmic complexity; and \citet{hu2025accelerating} accelerate generation via KV caching and autoregressive guidance. A complementary line studies ordering and complexity in diffusion LMs: \citet{feng2025theoretical} analyze sampling-step complexity under different metrics, and order-aware/adaptive inference procedures have been explored ~\cite{kim2025train}. Other related efforts examine training efficiency~\cite{ni2025superdatalearner} and connections between autoregressive and diffusion formulations~\cite{xue2025anyorder}. In contrast to these works that primarily modify \emph{decoding} or provide \emph{analysis}, we inject locality inductive bias during \emph{training} via locality-aware blockwise diffusion architectures to improve trainability.

\begin{figure*}[ht!]
\centering

\includegraphics[width=\textwidth, height=4.3cm]{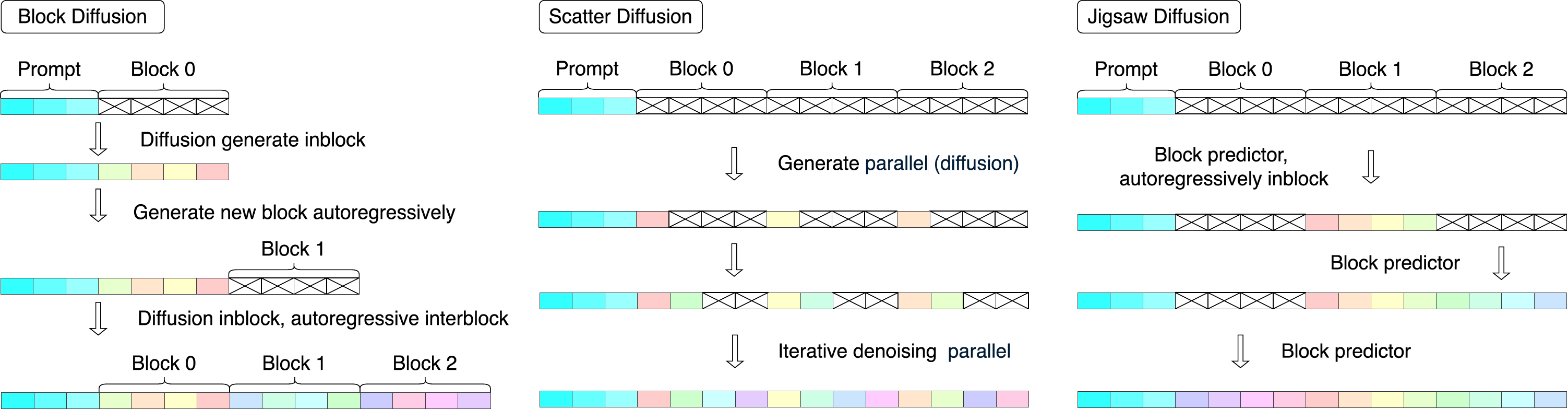}
\caption{Blockwise MDM variants. Block diffusion (left) \cite{arriola2025block}: Slides a generation window left-to-right; at each step, diffusion fills the tokens within the current window in parallel. Scatter diffusion (middle): Updates all blocks in parallel in a synchronized schedule, while generating tokens autoregressively within each block. Jigsaw diffusion (right): Updates blocks in a left-to-right order at the block level, while generating tokens autoregressively within each block, enabling iterative refinement across blocks.}
\label{fig:dllms-vs-ar-llms}
\end{figure*}

\section{Preliminaries}
\label{sect:preliminaries}

The goal of language modeling is to learn a parameterized distribution $p_\theta(\vx)$ that approximates the (unknown) data distribution $p_{\text{data}}(\vx)$. Let $\vx=(x_1,x_2,\ldots,x_L)$ denote a sequence drawn from $p_{\text{data}}(\vx)$. The model $p_\theta(\vx)$ assigns a probability to each possible sequence and is trained so that $p_\theta(\vx) \approx p_{\text{data}}(\vx)$. In practice, this is commonly achieved by minimizing the KL divergence $\operatorname{KL}(p_{\text{data}} \,\|\, p_\theta)$, which is equivalent to maximizing the expected log-likelihood:
\begin{equation}
\theta^* \in \argmax_{\theta} \; \E_{\vx \sim p_{\text{data}}} \left[ \log p_\theta(\vx) \right].
\label{equ:llm-obj}
\end{equation}

\noindent\textbf{AR-LLMs.}
Given a training corpus $\{\vx^1,\vx^2,\ldots,\vx^N\}$, autoregressive LLMs (AR-LLMs) factorize the likelihood of each sequence $\vx^j$ as
$p_\theta(\vx^j) = \prod_{i=1}^{n_j} p_\theta(x_i^j \mid x_{1:i-1}^j)$,
where by convention $p_\theta(x_1^j \mid x_{1:0}^j)=p_\theta(x_1^j)$.
The maximum-likelihood pretraining objective becomes
\begin{equation}
\theta^* \in \argmax_{\theta} \; \frac{1}{N} \sum_{j=1}^N \sum_{i=1}^{n_j} \log p_\theta(x_i^j \mid x_{1:i-1}^j).
\label{equ:ar-llm-obj}
\end{equation}
Most widely used LLMs \cite{radford2018improving,radford2019language,brown2020language,raffel2020exploring,touvron2023llama,bai2023qwen,team2023gemini,liu2024deepseek,zhao2023survey} follow this autoregressive training objective.

\noindent\textbf{Masked diffusion LMs (MDMs).}
Unlike AR-LLMs, which explicitly factorize $p_\theta(\vx)$ from left to right, masked diffusion language models define $p_\theta(\vx)$ implicitly by learning to \emph{reverse} a token corruption process.
Let $\mathcal{V}$ denote the vocabulary and let $M \notin \mathcal{V}$ be a special \texttt{[MASK]} token.
Given a clean training sequence $\vx^t=(x_1^t,\ldots,x_{n_t}^t)$, a standard masked-diffusion forward process samples a \emph{masking level} $\tau \sim \mathrm{U}(0,1)$ and independently corrupts each token:
\begin{align}
q_\tau(\vx_\tau^t \mid \vx^t) \;&=\; \prod_{i=1}^{n_t} q_\tau(x_{\tau,i}^t \mid x_i^t), \label{equ:dlm-forward} \\
q_\tau(x_{\tau,i}^t \mid x_i^t) \;&=\; (1-\tau)\,\mathbb{I}[x_{\tau,i}^t=x_i^t] + \tau\,\mathbb{I}[x_{\tau,i}^t=M], \nonumber
\end{align}
where $\vx_\tau^t=(x_{\tau,1}^t,\ldots,x_{\tau,n_t}^t)$ is the corrupted sequence.
An MDM parameterizes a \emph{mask predictor} $p_\theta(\cdot \mid \vx_\tau^t)$ that predicts the original token at each position conditioned on $\vx_\tau^t$ (typically producing all positions in parallel) corresponding to total $T$ denoising steps.
Training can be derived by minimizing a negative evidence lower bound (NELBO); for the absorbing \texttt{[MASK]} corruption in Eq.~\eqref{equ:dlm-forward}, this simplifies to a reweighted masked-token cross-entropy objective:
\begin{equation}
\hspace{-4mm}\theta^* \in \argmin_{\theta}\;
\frac{1}{N}\sum_{t=1}^N
\E_{\tau \sim \mathrm{U}(0,1),\, \vx_\tau^t \sim q_\tau(\cdot \mid \vx^t)}
\left[f_\tau(\vx_\tau^t) \right], \label{equ:dlm-obj}    
\end{equation}
\begin{equation}
f_\tau(\vx_\tau^t) := -\frac{1}{\tau}\sum_{i=1}^{n_t}
\mathbb{I}[x_{\tau,i}^t=M]\;\log p_\theta\!\left(x_i^t \mid \vx_\tau^t\right). \nonumber
\end{equation}
After training, generation simulates an approximate reverse process: it starts from a fully masked sequence (corresponding to $\tau \approx 1$) and iteratively fills in tokens as the masking level decreases toward $\tau \approx 0$. This yields non-autoregressive (parallel) updates within each denoising step.


\noindent\textbf{Block diffusion models.}
To bridge the gap between AR-LLMs and standard diffusion LMs, \emph{block diffusion} models, hereafter block diffusion (namely, BD3-LMs)  (or SDAR~\citealp{cheng2025sdar}), interpolate between these paradigms~\citep{arriola2025block}, as illustrated in Fig.~\ref{fig:dllms-vs-ar-llms} (left). This line of framework partitions a sequence $\vx$ into $K$ non-overlapping blocks,
$\vx = (\vx^{(1)}, \vx^{(2)}, \ldots, \vx^{(K)})$.
The joint distribution is factorized autoregressively at the block level, while tokens within each block are modeled via a conditional diffusion (denoising) process:
\begin{equation}
    p_\theta(\vx) = \prod_{k=1}^K p_\theta(\vx^{(k)} \mid \vx^{(<k)}),
    \label{equ:bd-factorization}
\end{equation}
where $\vx^{(<k)}$ denotes the concatenation of the preceding blocks.
Unlike ARs, where the conditional distribution is categorical over a single next token, here $p_\theta(\vx^{(k)} \mid \vx^{(<k)})$ represents a denoising distribution over an entire block.

During training, the conditioning context $\vx^{(<k)}$ remains clean (unmasked) and serves as a fixed prefix, while the target block $\vx^{(k)}$ undergoes the forward corruption process in Eq.~\eqref{equ:dlm-forward}.
The objective minimizes the sum of conditional NELBOs across blocks. For a single sequence $\vx$, this can be written as
\begin{equation}
\label{equ:bd-obj}
\mathcal{L}_{\text{BD}}(\theta)
= \sum_{k=1}^K \E_{\tau,\,\vx_\tau^{(k)}}
\Big[\,\ell_{\text{block}}(\vx^{(k)},\vx_\tau^{(k)},\vx^{(<k)})\,\Big],
\end{equation}
\begin{equation}
\ell_{\text{block}} := -\frac{1}{\tau}\sum_{x_j \in \vx^{(k)}} \mathbb{I}[x_{\tau,j}=M]\;
 \log p_\theta\!\left(x_j \mid \vx_\tau^{(k)}, \vx^{(<k)}\right). \nonumber
\end{equation}

This hybrid formulation offers two practical advantages.
First, because the prefix $\vx^{(<k)}$ is clean, the model can leverage key--value (KV) caching for the history, reducing compute compared to standard MDMs that re-encode the full sequence at every denoising step~\citep{cheng2025sdar}.
Second, the blockwise factorization introduces a coarse left-to-right inductive bias at the \emph{block} level. \citet{arriola2025block} show that this bias can stabilize training and improve perplexity on long sequences relative to fully non-autoregressive diffusion.
However, because generation within each block remains parallel, such models may still under-exploit \emph{token-level} locality, which is crucial for tasks with strict causal or left-to-right structure, a gap we address with our locality-aware blockwise diffusion architectures in the next sections. Motivated by this limitation, we reverse the roles of autoregression and diffusion: we enforce autoregressive generation \emph{within} each block to preserve token-level locality, while using diffusion \emph{across} blocks to retain iterative refinement at a coarser scale.

\section{Locality-aware Block Diffusion Models}
\label{sect:trained-models}

Standard block diffusion denoises tokens within each block in parallel, which can underutilize token-level left-to-right dependencies. In this section, we propose Scatter and Jigsaw, two locality-aware blockwise diffusion architectures that enforce autoregressive locality within blocks while preserving diffusion-style refinement across blocks.

Our goal is not to introduce a heavily engineered hybrid architecture, but to perform a \emph{controlled} intervention on generation order. 
Both Scatter and Jigsaw are intentionally minimal modifications built on the same backbone family, so that performance differences can be attributed primarily to factorization rather than architectural confounds. 
They are also motivated by two complementary principles. 
First, \textbf{Scatter} maximizes the spatial diversity of the available conditioning context at each intra-block offset, while preserving local causal structure within each block. 
Second, \textbf{Jigsaw} implements an ``easiest-first'' strategy by selecting the next block based on predictive entropy, greedily resolving low-uncertainty subproblems before harder ones. 
This design makes the two methods interpretable probes of how locality and planning interact with task dependency structure.


\subsection{Scatter: synchronized parallel autoregression}
\label{subsect:scatter}

Let the input training sequence $\vx \in \mathbb{R}^{L \times d}$ of length $L$ be partitioned into $K$ blocks of size $S$. At each step, the Scatter model generates the $j$-th token of \textit{all} blocks simultaneously in a synchronized manner, as illustrated in Fig.~\ref{fig:dllms-vs-ar-llms} (middle). That is, it adopts a column-major strategy, which decouples the generation order from the block order to maximize parallelism without sacrificing local causal dependencies. Even when the $j$-th tokens of the first two blocks are generated at the same step, their absolute positional indices are $j$ and $S+j$, respectively. This ensures the model retains distinct awareness of each block's global position and can implicitly learn inter-block dependencies via self-attention.

\textbf{Scatter causal mask.}  To enforce this ``synchronized'' logic during training, we design a \textit{Scatter Causal Mask} $\mathcal{M} \in \{0, 1\}^{L \times L}$. Let $u$ and $v$ be global indices mapping to blocks $k_u, k_v$ and intra-block offsets $j_u, j_v$ respectively. The mask is defined as 
\begin{equation}
\mathcal{M}_{u, v} = 
\begin{cases}
1 & \text{if } j_v < j_u \quad \text{(Earlier offset)}, \\
1 & \text{if } j_v = j_u \wedge k_v = k_u \quad \text{(Self)}, \\
0 & \text{otherwise}.
\end{cases}
\label{eq:scatter_mask}
\end{equation}
The above mask prevents a token from attending to any information at the same hierarchical level $j$ in other blocks (to ensure parallelism) or any future level $>j$.

In the generative phase, Scatter proceeds in $S$ macro-steps. At each macro-step $j \in [0, S-1]$, we target the set of indices $\mathcal{I}_j = \{j, S+j, 2S+j, \dots\}$. 
Unlike single-shot prediction, generation at each hierarchy level $j$ follows a standard iterative unmasking/denoising process. The model effectively solves $K$ parallel diffusion sub-problems, where each sub-problem is conditioned on the ``global past'' (indices with offset $<j$). This forces the model to resolve causal dependencies using the synchronized context.

\subsection{Jigsaw: entropy-guided dynamic planning}
\label{subsect:jigsaw}

While Scatter enforces strict synchronization, tasks like Sudoku or constraint satisfaction often benefit from non-monotonic generation, where ``easier'' or highly constrained regions are resolved first to reduce ambiguity for the remaining sequence. To enable this, we propose Jigsaw diffusion, which transforms generation into a dynamic planning process guided by model confidence. Specifically, let $\vx_{\text{context}}$ denote the set of currently determined tokens, initialized as the input prompt $\vx_{\text{prompt}}$, while the target response $\vx_{\text{response}}$ consists of the set of ungenerated blocks $\mathcal{U}$. The inference process iterates through two phases until completion:

\begin{enumerate}[leftmargin=*]
    \item \textbf{Uncertainty-guided planning:} We perform a non-destructive ``probe'' forward pass over all blocks in $\mathcal{U}$ to estimate token-wise predictive distributions $p_\theta(x_i | \vx_{\text{context}})$. We then quantify the uncertainty of each block $k$ via its average Shannon entropy:
    \begin{equation}
        \mathcal{H}(k) = \frac{1}{|S|} \sum_{i \in k} \mathcal{H}(p_\theta(x_i | \vx_{\text{context}})).
    \end{equation}
    The model identifies the easiest next step by greedily selecting the block with minimum entropy: $k^* = \argmin_{k \in \mathcal{U}} \mathcal{H}(k)$.

    \item \textbf{Localized generation:} The selected block $k^*$ is generated using standard autoregressive decoding (or fine-grained diffusion) conditioned strictly on $\vx_{\text{context}}$. Once resolved, $k^*$ is removed from $\mathcal{U}$ and appended to $\vx_{\text{context}}$, serving as a fixed anchor for subsequent iterations.
\end{enumerate}

\paragraph{Random block independence training.}
To support this arbitrary generation order, we train Jigsaw using a \textit{Random Independence Mask}. In this setup, target blocks in $\vx_{\text{response}}$ can attend to $\vx_{\text{context}}$ but are mutually invisible to each other. Within each target block, we enforce a local causal mask to align with the intra-block autoregressive decoding used during inference. This prevents the model from relying on spurious correlations from other ungenerated blocks, forcing it to reason solely based on the committed context.

\subsection{Complexity analysis}
\label{subsect:complexity}

We analyze inference complexity in terms of the number of function evaluations (NFEs), i.e., the number of model forward passes. Standard autoregressive models require $L$ forward passes to generate a sequence of length $L$. For MDMs, while inference admits a speed--quality trade-off, matching strong AR baselines typically requires fine-grained denoising schedules. State-of-the-art methods such as LLaDA \citep{nie2025large} and Dream \citep{ye2025dream} often employ strategies such as low-confidence remasking, for which the total number of diffusion steps $T$ can scale comparably to $L$. Thus, in the high-fidelity regime, the complexity of standard MDMs is also ${\mc O}(L)$. Under this high-fidelity assumption ($T \approx L$), we analyze our blockwise methods:

The generation of scatter diffusion is stratified into $S$ intra-block offsets (block size $S$). In a fully iterative setting, the cumulative number of denoising steps across all synchronized offsets sums to $L$. Consequently, Scatter has complexity ${\mc O}(L)$, matching standard diffusion baselines. While, the sequence of Jigsaw diffusion is generated as $K$ blocks ($K \times S = L$). For each block, the model performs one probe pass (entropy estimation) and $S$ solve passes (intra-block generation). The total number of NFEs is $K \times (1 + S) = K + L \approx L$ when $K \ll L$. Both Scatter and Jigsaw have linear complexity ${\mc O}(L)$, comparable to AR models and high-fidelity MDMs. Crucially, unlike AR models which are rigidly bound to $L$ steps, diffusion-style inference can be accelerated by using fewer sampling steps ($T \ll L$) when latency is prioritized over precision.

\section{Experiments}
\label{sect:experiments}

We evaluate AR, standard masked diffusion, Block diffusion, and our blockwise variants (Scatter, Jigsaw) on three controlled tasks (star-graph path-finding, in-context linear regression, and Sudoku solving) chosen to isolate (i) long-horizon planning, (ii) precise local feature binding, and (iii) global constraint satisfaction.

\textbf{Setups of model architectures.}  We consider two backbone types: (i) an AR baseline based on a \emph{standard decoder-only Transformer}, and (ii) MDMs based on a \emph{bidirectional Transformer} backbone. All models are \textbf{trained from scratch} with the same tokenizer, data, and training budget for each specific task, and are matched in parameter count across scales. Our blockwise diffusion variants share the same bidirectional backbone and identical parameter counts, and differ only in their method-specific attention masks / block schedules. 
We use RoPE~\cite{su2024roformer} across all models for positional encoding and do not initialize from any pre-trained checkpoints\footnote{See Appendix~\ref{app:model_architectures} for detailed architectural specifications.}.

\subsection{In-context linear regression}
\label{sec:icl-linear-regression}

\begin{figure*}[t]
\centering
\includegraphics[width=1.\textwidth]{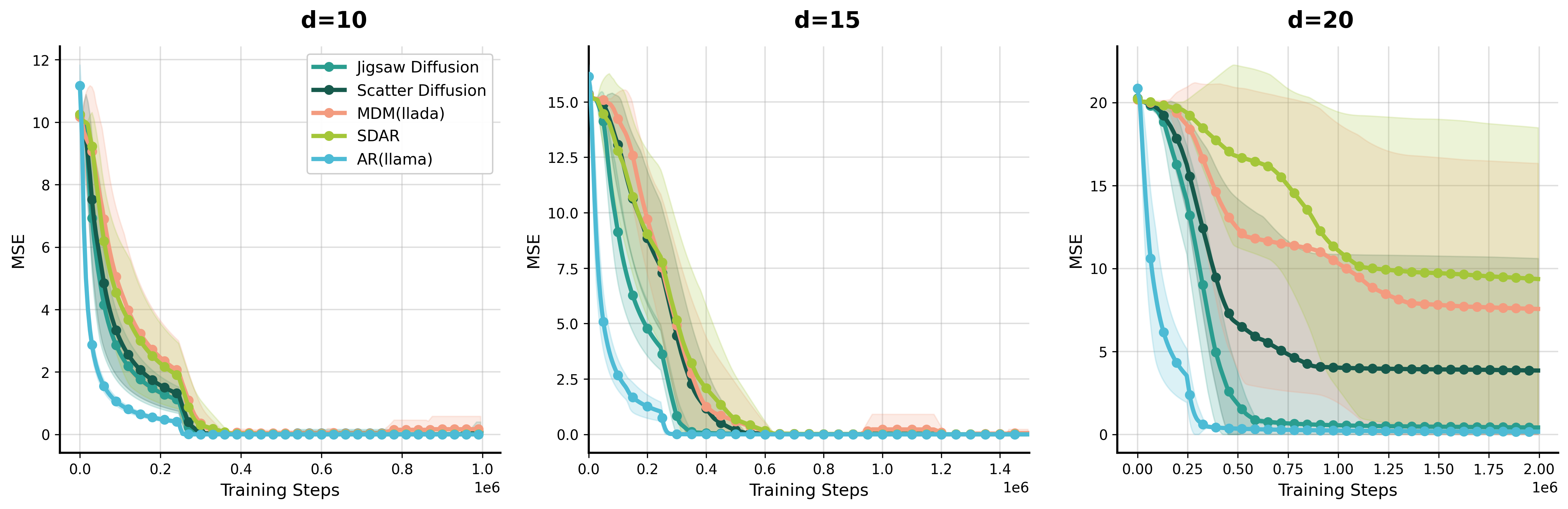} 
\caption{Validation MSE on in-context linear regression ($d \in \{10, 15, 20\}$). All models use parameter-matched templates (AR: LLaMA; MDM variants: LLaDA, see App.~\ref{app:model_architectures}). Lines: mean; shaded areas: $\pm 1$ SD. AR (light blue) and Jigsaw (teal) are the only paradigms to achieve near-zero MSE across all $d$. In contrast, MDM (orange) and SDAR (yellow-green) exhibit instability and high residual errors at $d=20$. Scatter Diffusion (dark green) converges rapidly but hits a performance floor at $d=20$, highlighting precision limitations of synchronized parallel denoising. \vspace{-2mm}}
\label{fig:icl_main_results}
\end{figure*}

\label{sec:icl_analysis}

To evaluate in-context learning and structural induction, we consider an in-context linear regression task based on linear function approximation \citep{Transformers_ICL}. 
In each sequence, the model must infer a latent linear operator $\vw$ from a set of demonstrations and then apply it to novel inputs within a single forward pass without parameter updates.

\textbf{Task formulation and results.} 
We define a function class $f(\vx) = \vw^\top \vx$, where the latent vector $\vw \sim \mathcal{N}({\bm 0}, \mI_d)$. 
For each example, inputs $\vx_i \sim \mathcal{N}({\bm 0}, \mI_d)$ are drawn i.i.d., and outputs are generated by $y_i = \vw^\top \vx_i$. 
We enforce a strict \textit{prompt-respond} structure to isolate contextual encoding from target:
\begin{equation}
\mc S = [\text{prompt context } \mc P] \parallel [\text{response targets } \mc R], \nonumber
\end{equation}
where the prompt context $\mc P$ contains $P$ demonstration pairs
$\mc P = [(\vx_1, y_1), \dots, (\vx_{P}, y_{P})]$,
and the response segment $\mc R$ contains $R$ query inputs for prediction,
$\mc R = [(\vx_{P+1}, \hat{y}_{P+1}), \dots, (\vx_{P+R}, \hat{y}_{P+R})]$,
with $\hat{y}_i$ denoting the model's predicted outputs. In our experiments, the sequence is constructed as a prefix of $P=20$ observable pairs $(\mathbf{x}_i, y_i)$, followed by $R=20$ query inputs where the model must reconstruct the corresponding target values. To evaluate scalability, we sweep across dimensions $d \in \{10, 15, 20\}$, employing unit-norm normalization in high-dimensional settings to force the model to capture directional mappings rather than magnitude-based shortcuts\footnote{See App.~\ref{app:icl_details_rigorous} for full mathematical specifications}. Fig.~\ref{fig:icl_main_results} reveals a capability gap that \textit{inverts} the results of the planning task:
\begin{equation}
\underbrace{\text{AR} \approx \text{Jigsaw}}_{\text{Success } (\text{MSE} \to 0)} \gg \underbrace{\text{Scatter}}_{\text{Partial}} \gg \underbrace{\text{SDAR} \approx \text{MDM}}_{\text{Failure } (\text{MSE} \approx d)}. \nonumber
\end{equation}

\textbf{Jigsaw matches AR stability on in-context linear regression task.} As shown in Fig.~\ref{fig:icl_main_results}, adding blockwise locality in Jigsaw ($S=4$) yields faster convergence and achieves structural recovery even at $d=20$. In contrast, other diffusion variants fail to recover the latent mapping reliably in this high-dimensional setting. Scatter improves quickly early on but plateaus at a \textit{performance floor} at $d=20$, indicating insufficient precision for high-dimensional feature binding. MDM and Block Diffusion exhibit pronounced optimization bottlenecks and training instability at $d=20$, and do not achieve structural recovery. These results suggest that purely global visibility (MDM) or overly rigid block-level causality (Block Diffusion) is ill-suited for tasks requiring precise sampling and label binding\footnote{Please find detailed analyses in App.~\ref{app:extended_icl_analysis}.}.

Comparing MDMs and their variants on the path-finding task (Section~\ref{sect:exp:path-finding}) and on linear regression reveals a striking \textit{paradigm inversion}:  global planning tasks (Sudoku, path-finding) benefit from holistic visibility (MDM), whereas local binding tasks (ICL) benefit from restricted locality (Jigsaw/AR). This suggests that generative architectures should be aligned with a task's intrinsic dependency structure, favoring global visibility for planning and stronger locality for precise input and output mapping.

\begin{figure*}[t]
\centering
\includegraphics[width=0.98\textwidth]{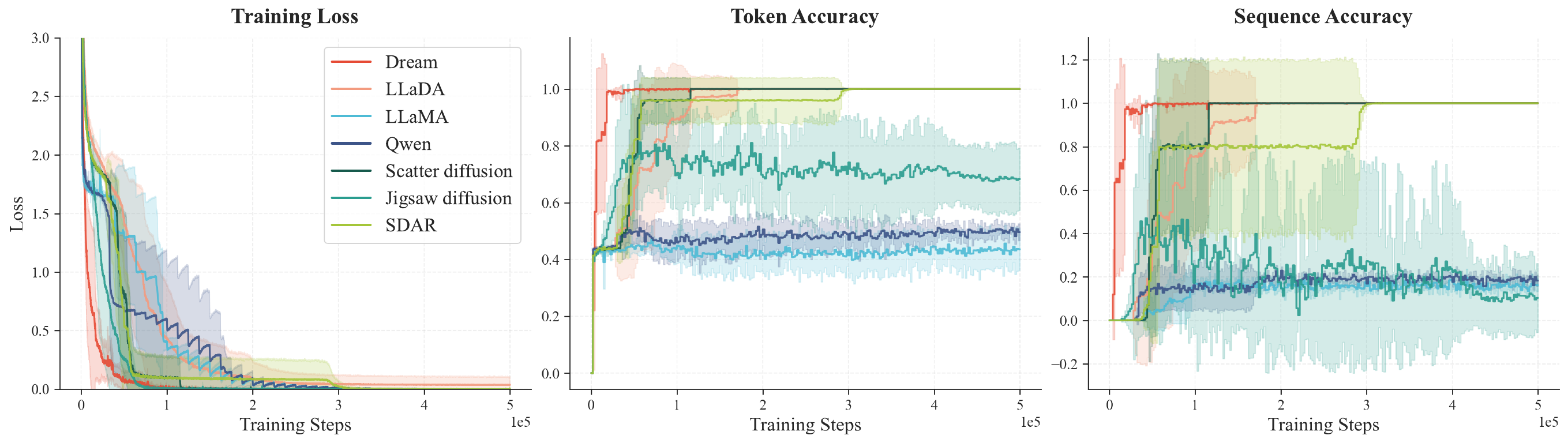}
\caption{Training dynamics on graph path-finding ($d=5, l=5$). \textit{Left}: Training loss. \textit{Middle}: Test token-match accuracy. \textit{Right}: Test sequence exact-match accuracy. Solid lines show the mean across 8 random seeds; shaded regions indicate one standard deviation. AR models plateau at random-guessing accuracy ($\approx 20\%$), and Jigsaw exhibits strong instability. In contrast, standard dLLM, Scatter, and Block Diffusion solve the task, reaching near-perfect accuracy.\vspace{-1mm}}
\label{fig:graph-results}
\end{figure*}

\subsection{Star-graph path-finding}
\label{sect:exp:path-finding}

\begin{figure}[H]
    \centering
    \includegraphics[width=0.45\textwidth]{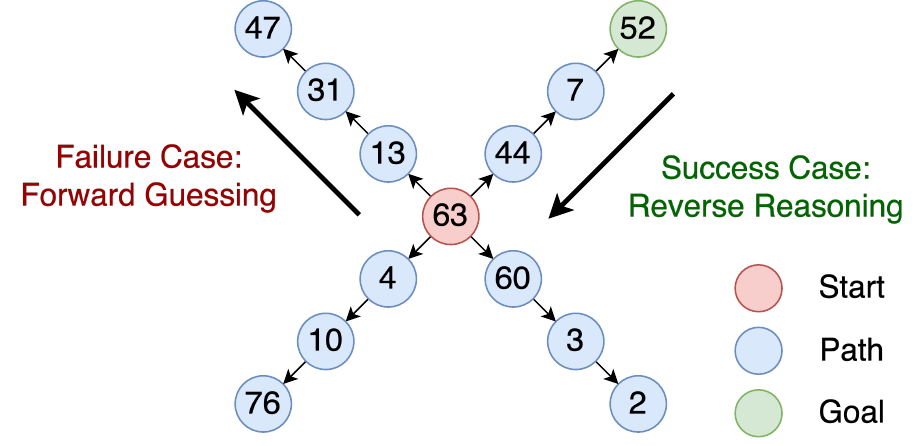}

    \caption{Illustration of the star-graph path-finding task. The ground truth path is $v_{\text{start}} = 63 \rightarrow 44 \rightarrow 7 \rightarrow 52 = v_{\text{goal}} $.}
    \label{fig:path-finding-task}
\end{figure}

We adopt the ``star-graph path-finding'' task \citep{bachmann2024pitfalls} to test lookahead planning. As shown in Fig.~\ref{fig:path-finding-task}, the logical dependency flows backwards (Goal $\to$ Start), which contradicts the forward generation order.

\textbf{Task formulation and results.} We construct synthetic graphs $G_{d,l}(N)$ where a central node $v_{start}$ has $d$ outgoing branches of length $l$, but only one branch leads to $v_{goal}$. 
We serialize the graph by shuffling all edges to remove positional cues, followed by a query prefix $\vp := \text{shuffled edge list} \;/\; v_{start}, v_{goal} =$. The target $\vr$ is the unique path from start to goal. To succeed, the model must implicitly trace the path backwards from $v_{goal}$ through the shuffled context; local transition heuristics from $v_{start}$ are insufficient (random-guessing accuracy is $1/d$). We report sequence exact match accuracy as in Fig.~\ref{fig:graph-results}. The results reveals a distinct performance hierarchy capability gap:

\begin{equation}
\underbrace{\text{dLLM} \approx \text{Scatter} \approx \text{Block}}_{\text{Success } (\approx 1.0)} \gg \underbrace{\text{Jigsaw} \approx \text{AR}}_{\text{Failure } (\approx 0.2)}.
\end{equation}

\textbf{MDMs overcome the lookahead barrier.} The AR baseline exhibits the classic ``Clever Hans'' failure \citep{bachmann2024pitfalls}: although training loss converges (memorization), test accuracy stagnates at $1/d$, consistent with random guessing. In contrast, \textit{standard MDMs, Scatter, and Block diffusion methods achieve near-perfect accuracy ($\approx 100\%$).} By treating generation as global denoising, these models can propagate constraints from $v_{goal}$ back to $v_{start}$; moreover, Scatter and Block retain this capability despite their blockwise left-to-right bias that corrects early-stage errors as partially illustrated in Table \ref{tab:denoising-log}.

\begin{table}[ht]
\centering
\footnotesize
\setlength{\tabcolsep}{6pt}
\caption{Denoising trajectories for path-finding ($d=5$). The successful run of Scatter implicitly learns to denoise from \textit{right-to-left}, whereas the failed run mimics AR-style \textit{left-to-right} guessing.}
\label{tab:denoising-log}
\begin{tabularx}{\columnwidth}{@{}c c l X@{}}
\toprule
\textbf{Step} & \textbf{Sequence Snapshot} & \textbf{Behavior} \\
\midrule
0 & \texttt{\$, \$, \$, \$, \$} & All Masked \\
2 & \texttt{67, \$, \$, \$, 72} & Anchors predicted \\
\midrule
\textbf{3} (\textcolor{red}{\xmark}) & \texttt{67, \textcolor{red}{\textbf{82}}, \dots, 72} &  Forward (AR) \\
\textbf{3} (\textcolor{teal}{\checkmark}) & \texttt{67, \dots, \textcolor{teal}{\textbf{9}}, 72} & Reverse (Scatter) \\
\midrule
\textbf{5} (\textcolor{red}{\xmark}) & \texttt{67, \textcolor{red}{\textbf{82, 54, 5}}, 72} & Incoherent Path (AR) \\
\textbf{5} (\textcolor{teal}{\checkmark}) & \texttt{67, \textcolor{teal}{\textbf{26, 51, 9}}, 72} & Exact Match (Scatter) \\
\bottomrule
\end{tabularx}
\end{table}

\textbf{Jigsaw has structural conflicts.} Unexpectedly, Jigsaw performs poorly, mirroring the AR baseline. We attribute this to a \textit{structural conflict} between the macro-level selection policy and the micro-level decoding mechanism. While Jigsaw's macro-policy correctly identifies goal-adjacent blocks as low-entropy (high-confidence), its \textit{intra-block} autoregressive decoding enforces a strict left-to-right order. As a result, the first token in a selected block must be predicted using only the currently committed prefix; lacking informative local context, this prediction collapses to AR-like random guessing, effectively severing the reverse logical chain.
This diagnosis is empirically validated by our ablation study, \footnote{See Appendix~\ref{subsec:graph_bs_ablation} for detailed analyses}, where reducing the block size to 1 eliminates the intra-block bottleneck and restores perfect performance.

\begin{figure}[t]
    \centering
    \includegraphics[width=0.5\textwidth]{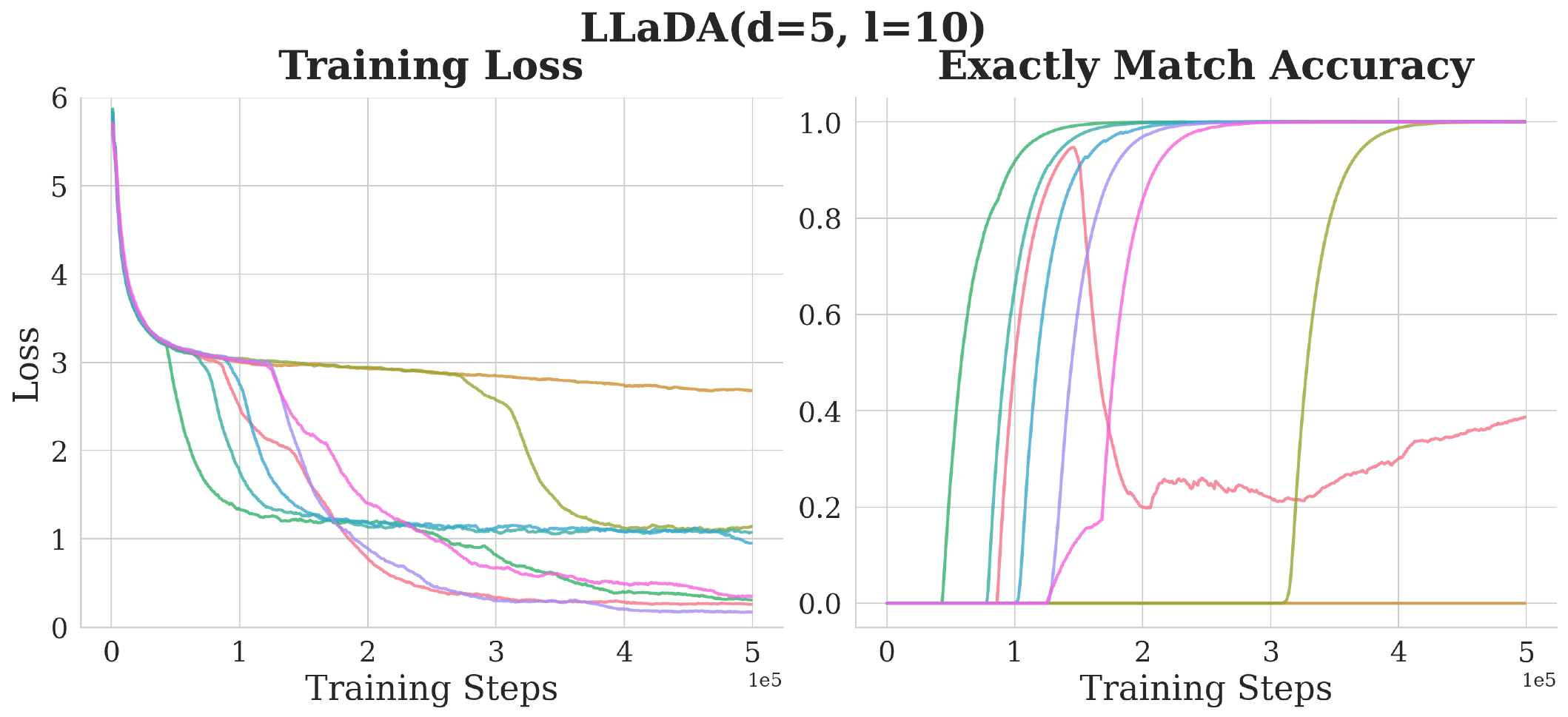}
    \caption{The training loss and accuracy as a function of training steps for 8 independent runs with different random seeds on standard MDM. The divergence across runs where some succeed while others stagnate, highlights sensitivity to initialization and the stochasticity of discovering the reverse-planning trajectory.\vspace{-2mm}}
    \label{fig:llada_results}
\end{figure}

\textbf{MDMs exhibit significant training instability.} 
Despite their strong accuracy, diffusion models exhibit significant training instability (Fig.~\ref{fig:llada_results}). Standard diffusion training uses \textit{isotropic} random masking, covering all generation orders. However, the task is highly \textit{anisotropic}: only \textit{reverse-order} trajectories provide a robust learning signal. To investigate whether modifying the masking distribution alone can alleviate this issue, we evaluated directionally-biased non-uniform masking schedules. However, we found this to be insufficient without explicit architectural inductive biases (see Appendix~\ref{app:non_uniform_masking}).
This phenomenon aligns with the ``subgoal imbalance'' identified by \citet{ye2025beyond}, where certain generation steps (e.g., forward lookahead) are intrinsically harder than others. Training therefore becomes a stochastic search: seeds that fail to discover the reverse trajectory early can collapse into an AR-like forward-guessing mode. This highlights the need for inductive biases that prune the search space.

\subsection{Sudoku solving}
\label{sect:exp:sudoku}

Sudoku serves as a probe for \textit{global constraint satisfaction}. The task requires completing a $9 \times 9$ grid governed by dense row, column, and $3 \times 3$ block mutual exclusion rules. To restore the 2D spatial topology lost during 1D serialization, we implement \textit{coordinate awareness} as a topological inductive bias (see App.~\ref{app:sudoku_details_rigorous} for math formalization).

\begin{figure}[t]
\centering
\includegraphics[width=\linewidth]{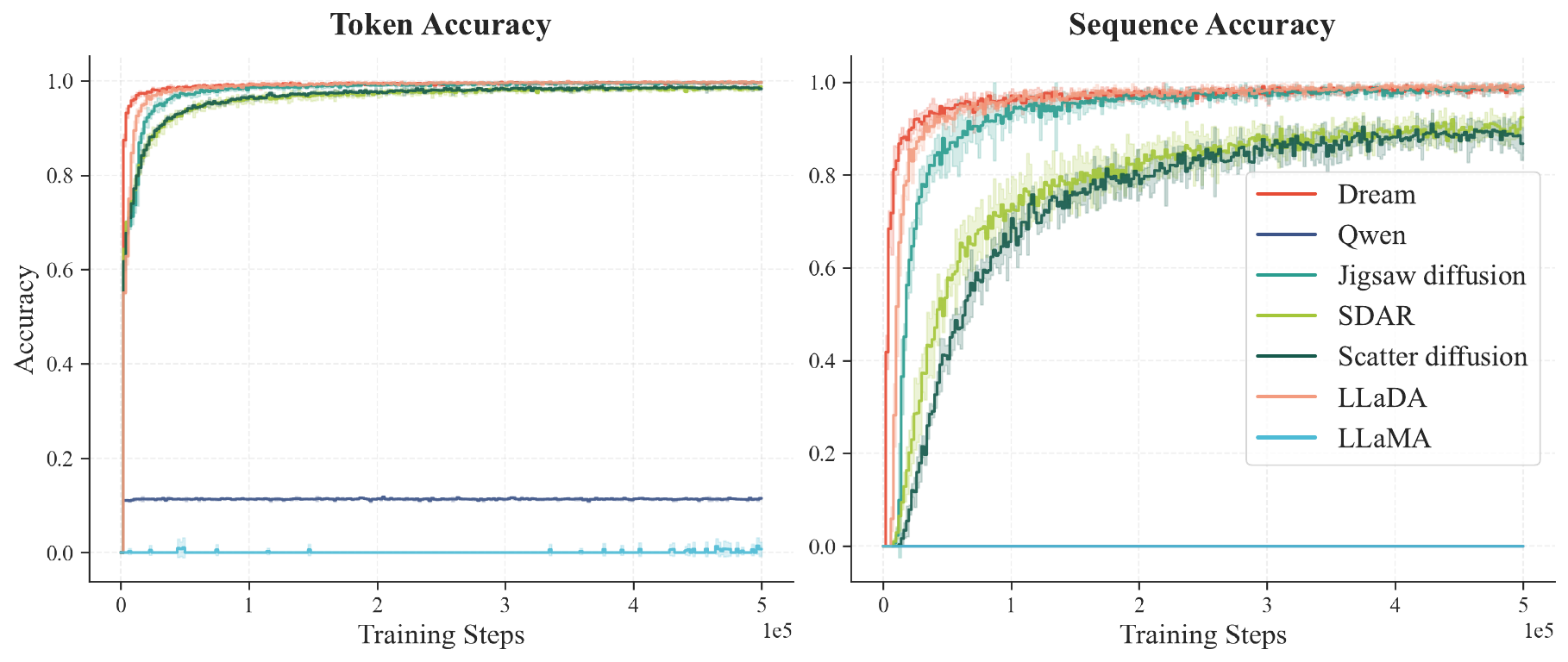} 
\caption{\textbf{Accuracy on Sudoku validation set.} Models with bidirectional visibility (LLaDA-based and Dream based MDMs) and \textbf{Jigsaw} achieve near-perfect solving rates. The \textbf{AR baseline} (LLaMA-based and Qwen-based) fails catastrophically ($0\%$ accuracy). All models are trained from scratch with parameter parity. }
\label{fig:sudoku_results}
\end{figure}

As illustrated in Fig.~\ref{fig:sudoku_results}, the Sudoku task produces the most polarized results among our experiments, yielding a hierarchy that mirrors path-finding but inverts the ICL results 
\begin{equation}
\underbrace{\text{MDM} \approx \text{Jigsaw}}_{\text{Success } (\approx 100\%)} \gg \underbrace{\text{SDAR} > \text{Scatter}}_{\text{Partial Success}} \ggg \underbrace{\text{AR}}_{\text{Failure } (0.0)}. \nonumber
\end{equation}


\textbf{The factorization curse in Sudoku puzzles.} Sudoku poses a fundamental challenge to AR models due to what we term the \textit{factorization curse}. Because Sudoku constraints are inherently non-causal and multi-directional, factorizing the joint distribution $p(\vx_{\text{res}} \mid \vx_{\text{prompt}})$ into a strict left-to-right chain of conditionals $\prod_{i} p(x_i \mid x_{<i}, \vx_{\text{prompt}})$ is misaligned with the problem structure. In an AR trajectory, a single early cell error can trigger a cascade of constraint violations that cannot be retrospectively corrected, since generation lacks a global revision mechanism. This failure mode is reflected empirically: AR models (LLaMA template) remain at $0\%$ accuracy throughout training. In contrast, diffusion paradigms (LLaDA template) perform iterative refinement, allowing the model to resolve high-certainty dependencies in a flexible order.


\textbf{Locality--consistency trade-off.} Among blockwise variants, SDAR and Scatter ($\approx 85\%$) exhibit lower ceilings, suggesting a trade-off between enforcing locality and maintaining global consistency. While SDAR uses bidirectional attention within rows, its fixed row-major order imposes a semi-causal bottleneck that hinders column-wise constraint propagation. Scatter’s hierarchical generation, which benefits token-local binding in linear regression, struggles with the tight horizontal--vertical coupling required for globally valid Sudoku solutions, leading to lower puzzle-level accuracy.

\subsection{Language modeling on LM1B}
\label{sec:exp:lm1b}
We further evaluate our method on the LM1B benchmark ($\sim$32B tokens) to test whether the proposed generation order transfers to realistic language modeling. 
This experiment is intended as a controlled validation of the factorization bias studied in this paper, rather than a large-scale language-model scaling study. 
For comparison, we use Block Diffusion Language Models (BD3-LMs) as a matched baseline.

\textbf{Setup.}
For a controlled comparison, we implement \textbf{Scatter Diffusion} within the official BD3-LMs codebase~\citep{arriola2025block}. 
We keep the model size ($110$M parameters), tokenizer (\texttt{bert-base-uncased}), context length ($128$), optimizer, and all other training hyperparameters identical to the BD3-LMs setup; the only change is the generation order induced by Scatter. 
We compare final test perplexity (PPL) for block sizes $L \in \{4,8,16\}$. 
We focus on Scatter here because it can be incorporated into the BD3-LMs codebase without changing the backbone or training recipe. 
In contrast, Jigsaw uses entropy-guided block selection and autoregressive intra-block decoding, so a fair comparison within the same codebase would require additional architectural changes beyond generation order.

\begin{table}[t]
\centering
\caption{Test perplexity on LM1B ($\sim$32B tokens). Lower is better. ``Reported'' denotes numbers from the BD3-LMs paper; ``Ours'' denotes our runs under the matched codebase setting.}
\label{tab:lm1b_main}
\small
\begin{tabular}{llcc}
\toprule
Category & $L$ & BD3-LMs & Scatter Diffusion \\
\midrule
Reported & 4  & 28.23 & -- \\
Reported & 8  & 30.60 & -- \\
Reported & 16 & 29.83 & -- \\
\midrule
Ours & 4  & 28.20 & 32.10 \\
Ours & 8  & 29.70 & 30.50 \\
Ours & 16 & 32.90 & \textbf{27.10} \\
\bottomrule
\end{tabular}
\vspace{-2mm}
\end{table}

\textbf{Results.}
As shown in Table~\ref{tab:lm1b_main}, Scatter Diffusion achieves a test PPL of \textbf{27.10} at $L=16$, which is competitive with---and numerically better than---the best reported BD3-LMs result (28.23 at $L=4$). 
This result provides direct evidence that our proposed generation order is not limited to synthetic settings: when instantiated in a standard language-modeling codebase with matched architecture and hyperparameters, Scatter remains capable of modeling realistic natural language data.

\textbf{Scaling evidence.}
To further test whether the blockwise inductive bias scales with model size, we additionally train a medium-scale Scatter model (hidden size $1024$, $24$ layers, $\sim 350$M parameters). 
The larger model consistently improves over the $110$M Scatter baseline and achieves the best result at $L=16$.

\begin{table}[t]
\centering
\caption{Scaling behavior of Scatter Diffusion on LM1B. Lower is better. BD3-LMs only reports $110$M-scale results on LM1B in their paper.}
\label{tab:lm1b_scale}
\small
\begin{tabular}{lccc}
\toprule
$L$ & BD3-LMs (110M) & Scatter (110M) & Scatter (350M) \\
\midrule
4  & 28.23 & 32.10 & 31.63 \\
8  & 30.60 & 30.50 & 30.02 \\
16 & 29.83 & 27.10 & \textbf{26.58} \\
\bottomrule
\end{tabular}
\vspace{-2mm}
\end{table}

The medium-scale Scatter model reaches a test PPL of \textbf{26.58} at $L=16$, improving over both the $110$M Scatter model (27.10) and the best BD3-LMs result reported in the original paper (28.23). 
This suggests that the locality-aware blockwise inductive bias scales positively with model capacity, at least in this initial LM1B setting.

\textbf{Takeaway.}
These LM1B results complement our controlled synthetic experiments. 
The synthetic tasks let us causally isolate how generation order interacts with dependency structure, while LM1B confirms that the same factorization design can transfer to realistic language modeling. 
Together, they support our central claim that generation order is not merely an implementation detail, but a meaningful inductive bias for diffusion LMs.


\section{Discussion and Conclusion}

\textbf{Discussion.} Our results show that masked diffusion language models (MDMs) and autoregressive LLMs (AR-LLMs) excel on different dependency structures. On global refinement and constraint satisfaction (Sudoku) and reverse-dependency planning (star-graph path-finding), diffusion-style iterative denoising enables isotropic information flow and revision, which can overcome the limitations of a fixed left-to-right factorization. In contrast, on tasks with strict prompt-to-response binding and strong token-local structure (in-context linear regression), standard random-masking MDMs are substantially less stable, exhibiting high variance across seeds and frequent optimization failures.

A key implication is that \emph{isotropic random masking is likely a suboptimal inductive bias for ordered generation}. When only a small subset of generation trajectories provides a strong learning signal, training under uniform random masking effectively forces the model to average over many incompatible orders, turning optimization into a stochastic search. This helps explain why MDMs can succeed yet remain fragile on anisotropic tasks, and why blockwise diffusion that still denoises tokens in parallel may under-exploit the token-local left-to-right dependencies that make AR training robust.

Our locality-aware blockwise designs (Scatter and Jigsaw) mitigate these issues by injecting left-to-right locality at the right granularity while preserving diffusion-style refinement across blocks. Empirically, they improve trainability and reduce variance on ordered tasks (linear regression) while retaining strong performance on globally constrained tasks (Sudoku), indicating that \emph{locality is a key ingredient for stable MDM training}. More broadly, these findings motivate diffusion LMs beyond random masking, including designs that better align the corruption process with task structure and potentially leverage longer or multi-scale vocabularies so that denoising operates over locally coherent units rather than overly fine-grained tokens.

\textbf{Conclusion.} We compared AR-LLMs and MDMs on three controlled structured-generation tasks and found a consistent task-dependent trade-off: diffusion benefits planning and global constraint satisfaction, while AR-style locality benefits precise feature binding. To bridge this gap, we introduced Scatter and Jigsaw, which enforce token-local autoregressive structure within blocks while enabling diffusion-style refinement across blocks; these methods improve stability and compute-to-target on strongly ordered tasks and remain competitive on Sudoku. Overall, our results suggest that standard random-masking MDMs may not be an optimal diffusion LM design for ordered generation, and that locality-aware (and potentially multi-scale) diffusion architectures are a promising direction for more stable and general diffusion language modeling.

\clearpage
\section*{Impact Statement}
\label{sect:impact}
This paper presents work whose goal is to advance the field of Machine Learning. There are many potential societal consequences of our work, none of which we feel must be specifically highlighted here.

\bibliography{references}
\bibliographystyle{icml2026}

\newpage
\appendix
\onecolumn

\newpage
\appendix
\onecolumn

\section{Model Architecture and Training Configurations}
\label{app:model_architectures}

To ensure a rigorous comparison, we standardize the underlying neural network parameters (e.g., embedding dimensions, layers, and heads) across all scales. Crucially, all models are trained from scratch without pre-trained weights. We use "LLaMA-based" and "LLaDA-based" to refer to architectural templates (RMSNorm, SwiGLU, and RoPE) and masking strategies, rather than specific checkpoints.

\subsection{Backbone Templates and Parity}

Within each scale (Small, Medium, Large), we enforce strict parameter parity. An AR model and its MDM counterpart share the same $n_{embd}$, $n_{layer}$, and $n_{head}$, ensuring that performance gaps stem from the learning paradigm rather than model capacity.

\paragraph{Causal Template (AR-LLMs).}
Our autoregressive models follow the LLaMA/Qwen architectural template\cite{touvron2023llama,bai2023qwen}. These are decoder-only Transformers utilizing a \textbf{strict lower-triangular causal attention mask}. They are trained via next-token prediction and employ Key-Value (KV) caching for efficient scalar generation during inference.

\paragraph{Bidirectional Template (MDMs and Blockwise Variants).}
For the masked diffusion and locality-aware blockwise paradigms—specifically standard MDM (following LLaDA \citealp{nie2025large} and Dream \citealp{ye2025dream}), as well as our proposed Scatter, Jigsaw, and the SDAR baseline—we utilize the LLaDA Transformer backbone. 
\begin{itemize}[leftmargin=*]
    \item Bidirectional Visibility: Unlike the causal template, the LLaDA-based foundation is inherently bidirectional, allowing tokens to attend to all masked/unmasked positions in the current sequence.
    \item Custom Attention Masks: While sharing identical bidirectional parameters, we inject specialized attention masks ($\mathcal{M}$) into the LLaDA backbone to implement different locality constraints (e.g., Scatter's synchronized offsets or Jigsaw's random independence) without losing the ability to perform parallel denoising.
\end{itemize}





\subsection{Implementation Details of Generative Paradigms}
\label{app:impl_details}

We categorize the evaluated models into three paradigms based on their architectural templates and training objectives. All variants share the same underlying parameter scale (Small, Medium, or Large) as defined in Table~\ref{tab:model_configs_app}.

\paragraph{Autoregressive LLMs (AR-LLMs).}
These models follow the LLaMA \cite{touvron2023llama} and Qwen \cite{bai2023qwen} architectural templates, utilizing a decoder-only Transformer with a strict lower-triangular causal attention mask. The training objective is standard cross-entropy loss over next-token prediction.

\paragraph{Masked Diffusion Models (MDMs).}
The standard MDM baselines follow the LLaDA \cite{nie2025large} and Dream \cite{ye2025dream} style, using a bidirectional Transformer backbone. They are trained via the reweighted masked-token cross-entropy objective as defined in Eq.~\ref{equ:dlm-obj}.

\paragraph{Locality-Aware Blockwise Variants.}
Our proposed models (Scatter, Jigsaw) and the SDAR baseline leverage the same LLaDA-based bidirectional backbone as MDMs, but utilize method-specific attention masks ($\mathcal{M}$) during training and inference:

\begin{itemize}[leftmargin=*]
    \item SDAR: Employs a block-causal mask where tokens within a block are bidirectional but only attend to previously generated blocks.
    \item Scatter (Ours): Implements a scatter causal mask to support synchronized parallel autoregression across predefined block offsets.
    \item Jigsaw (Ours): Utilizes a random independence mask during training, where target blocks are mutually invisible to support uncertainty-guided dynamic planning.
\end{itemize}

\subsection{Complexity comparison}
Table \ref{tab:complexity} summarizes the Number of Function Evaluations (NFEs). In our high-fidelity benchmarking, we assume $T \approx L$ for MDMs.

\begin{table}[ht]
\centering
\caption{Inference Complexity Analysis. $L$: length, $S$: block size, $K$: blocks ($L = K \times S$).}
\label{tab:complexity}
\begin{small}
\begin{tabular}{lcll}
\toprule
\textbf{Paradigm} & \textbf{Backbone} & \textbf{NFEs} & \textbf{Factorization Principle} \\ \midrule
AR & LLaMA & $L$ & Scalar Sequential (Fixed) \\
MDM & LLaDA / Dream & $T \approx L$ & Holistic (No fixed order) \\
Block (SDAR) & LLaDA-based & $K \times T_{block}$ & Block-Sequential \\
Scatter (Ours) & LLaDA-based & $S \times T_{offset}$ & Offset-Sequential (Synchronized) \\
Jigsaw (Ours) & LLaDA-based & $K \times (1 + S)$ & Dynamic Planning (Non-monotonic) \\ \bottomrule
\end{tabular}
\end{small}
\end{table}

\subsection{Model scale and hyperparameters}
All paradigms are evaluated across three scales, with parameter parity maintained between LLaMA and LLaDA-based backbones.

\begin{table}[ht]
\centering
\caption{Model Scale Configurations. Hyperparameters ($n_{embd}, n_{layer}, n_{head}$) are shared by all generative paradigms within the same task category. ICL and Path Finding (PF) tasks utilize a deeper architectural configuration, while Sudoku follows a wider but shallower setup.}
\label{tab:model_configs_app}
\begin{tabular}{lcccc}
\toprule
\textbf{Configuration} & \textbf{$n_{embd}$} & \textbf{$n_{layer}$} & \textbf{$n_{head}$} & \textbf{Approx. Params} \\ \midrule
Small (ICL \& Path-finding / Sudoku)     & 192 / 384 & 8 / 6  & 6  & $\sim$4M / 12M   \\
Medium (ICL \&  Path-finding / Sudoku)    & 256 / 512 & 12 / 8 & 8  & $\sim$10M / 30M  \\
Large (ICL \&  Path-finding / Sudoku)     & 384 / 640 & 16 / 10 & 12 / 10 & $\sim$30M / 60M  \\ \bottomrule
\end{tabular}
\end{table}

\section{Experimental Details}
\label{app:exp-details}

We provide a rigorous summary of the experimental configurations. To ensure a controlled comparison, all five paradigms—\textbf{AR, MDM, Jigsaw, Scatter, and SDAR}—utilize a unified Transformer-based backbone (GPT-2 style) with task-specific modifications to attention masks and positional encodings.

\subsection{In-Context linear regression (ICL)}
\label{app:icl_details_rigorous}

\paragraph{Task Formulation.}  The ICL task probes \textbf{structural induction}—the ability to extract a latent operator $\vw$ from demonstrations. We define a function class $f(\vx) = \vw^\top \vx$, where the latent vector $\vw \sim \mathcal{N}(0, \mI_d)$ and query inputs $\vx_i \sim \mathcal{N}(0, \mI_d)$ are sampled from an isotropic Gaussian prior. The model must learn to perform functional mapping within a single forward pass without parameter updates.

\paragraph{Binding-Oriented Serialization.} Unlike standard sequence modeling, ICL requires precise binding between $\vx$ and $y$. We enforce a strict Prompt--Respond (P-R) structure to isolate contextual encoding from target decoding: 
\begin{equation}
    \mathcal{S} = [ \underbrace{(\vx_1, y_1), \dots, (\vx_{P}, y_{P})}_{\text{Prompt Context } \mathcal{P}} ] \parallel [ \underbrace{(\vx_{P+1}, \hat{y}_{P+1}), \dots, (\vx_{P+R}, \hat{y}_{P+R})}_{\text{Respond Target } \mathcal{R}} ],
\end{equation}
where $P=20$ and $R=20$. Here, $P$ (Prompt) represents the number of demonstration pairs where both $\vx$ and the ground-truth $y$ are fully observable, serving as the context for inducing $\vw$. $R$ (Respond) represents the query targets where the model is provided with $\vx$ and must predict the response $\hat{y}$ (i.e., $y$ is corrupted or masked during training). This partition ensures the model performs functional induction from $\mathcal{P}$ rather than simple sequence memorization.




\paragraph{Evaluation Metrics and Training Budget.} 
Success is quantified by the Respond MSE over the $R$ targets: 
\begin{equation}
    \mathcal{L}_{\text{MSE}} = \mathbb{E}_{\mathcal{S} \sim \mathcal{D }} \left[ \frac{1}{R} \sum_{j=P+1}^{P+R} \| \hat{y}_{j} - y_{j} \|^2 \right],
\end{equation}
calculated exclusively over the respond region to isolate predictive precision. The validation set consists of 3,200 held-out task instances. To ensure a stable estimate of structural recovery, the validation MSE is computed across 50 batches with a batch size of 64. We track performance across training iterations scaled by the task complexity: $1 \times 10^6$, $1.5 \times 10^6$, and $2 \times 10^6$ steps for $d=10, 15, 20$ respectively, ensuring sufficient budget to capture the structural transition.

\begin{table}[ht]
\centering
\caption{\textbf{Key Parameters for ICL Structural Induction.}}
\label{tab:icl_key_params}
\begin{tabular}{lll}
\toprule
\textbf{Parameter} & \textbf{Value} & \textbf{Rational/Source} \\ \midrule
$d$ (Dimension) & $\{10, 15, 20\}$ & Task complexity scaling \\
$S$ (Block size) & $\{2, 4, 10\}$ & Locality vs. context balance ablation \\
Batch Size & $64$ & Optimization stability \\
Steps & $1\text{M} \to 2\text{M}$ & Capturing the "epiphany" phase \\
$\eta$ (Learning Rate) & $2 \times 10^{-4}$ & Constant rate for stable induction \\
\bottomrule
\end{tabular}
\end{table}





\subsection{Extended Analysis of ICL Performance}
\label{app:extended_icl_analysis}

\begin{figure*}[ht]
    \centering
    \includegraphics[width=1\textwidth]{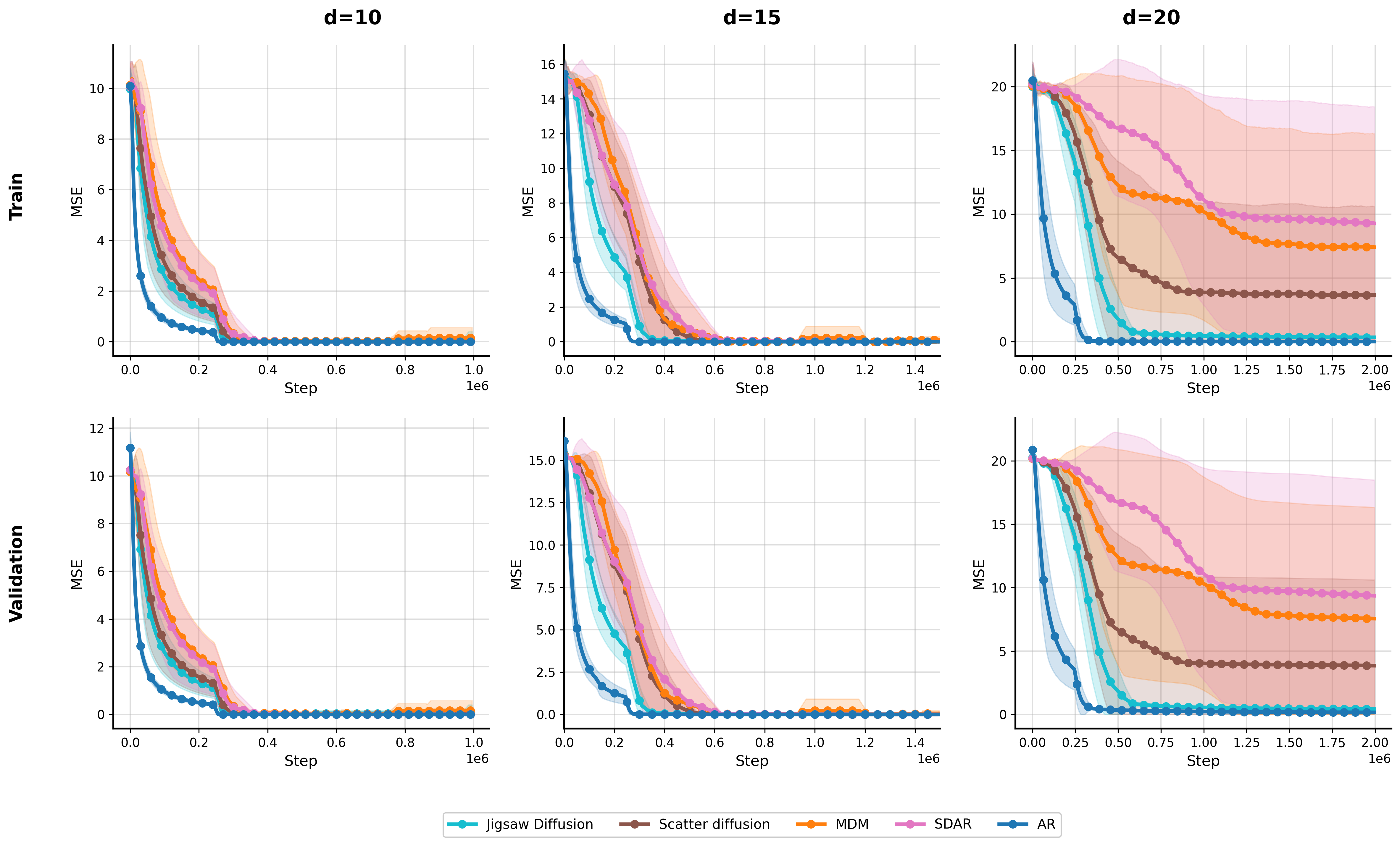} 
    \caption{Learning trajectories for In-Context Linear Regression across dimensions $d$. \textbf{Models follow parameter-matched LLaMA (AR) and LLaDA (MDM/blockwise) templates.} The top and bottom rows show Train MSE and \textbf{Validation MSE}, respectively. While all paradigms perform competitively in low-D regimes ($d=10$), a clear performance hierarchy emerges as $d$ increases. AR (dark blue) and Jigsaw Diffusion (cyan) are the only paradigms that consistently achieve near-zero MSE in high-D settings ($d=20$), whereas MDM (orange) and SDAR (pink) struggle with optimization plateaus.}
    \label{fig:icl_full_results}
\end{figure*}

In this section, we provide a comprehensive analysis of the learning trajectories across varying problem dimensions $d \in \{10, 15, 20\}$, as visualized in Fig. \ref{fig:icl_full_results}. This detailed view reveals the transition from statistical mean-prediction to precise structural induction.


\paragraph{Numerical Evidence of Mean-Prediction.} 
A key observation from Fig. \ref{fig:icl_full_results} is the starting point of the MSE curves at Step 0. For each dimension $d$, all models—regardless of their generative paradigm—commence with an $\text{MSE} \approx d$ (e.g., $\text{MSE} \approx 20$ for $d=20$). From a statistical perspective, for Gaussian inputs $x \sim \mathcal{N}(0, \mathbf{I})$ and a unit-variance operator $\mathbf{w}$, the MSE of a zero-predictor or a mean-predictor is exactly $d$. This empirically confirms that before the "structural epiphany" (the sudden drop in MSE), models default to predicting the mean of the distribution ($y=0$), which is the optimal risk-minimization strategy in the absence of induced latent structure.

\paragraph{The $S=4$ Watershed and Jigsaw's Efficiency.} 
As the complexity increases to $d=20$, Jigsaw Diffusion (cyan) demonstrates remarkable robustness, significantly outperforming other diffusion variants and trailing only the AR baseline. We attribute this to the $S=4$ Watershed mechanism discussed in App. \ref{subsec:icl_bs_ablation}. By utilizing an intra-block AR mask within a small window ($S=4$), Jigsaw isolates a manageable number of $(x, y)$ pairs, allowing the bidirectional backbone to perform interference-free feature binding. This allows Jigsaw to "exit" the mean-prediction plateau far earlier than standard MDMs.

\paragraph{Performance Floor and Instability in High-D Regimes.}
The "performance hierarchy" becomes most visible at $d=20$:
\begin{itemize}
    \item \textbf{Scatter Diffusion (Brown):} Hits a performance floor around $\text{MSE} \approx 4$. Its synchronized parallel nature, while efficient, forces the model to predict all tokens in a block simultaneously without the benefit of the fine-grained, step-by-step revision provided by Jigsaw's intra-block causal decoding.
    \item \textbf{MDM (Orange) and SDAR (Pink):} These paradigms exhibit significant instability and high variance (indicated by the wide shaded regions) in the $d=20$ regime. Their trajectories often stagnate at high MSE levels ($10$--$15$), failing to achieve structural induction. This optimization failure is likely due to destructive interference: in a fully visible bidirectional window, the model is overwhelmed by the $d^2$ complexity of the high-dimensional operator, where gradient signals from different prompt pairs effectively cancel each other out.
\end{itemize}
These results reinforce the conclusion that for algebraic tasks requiring high precision, \textbf{local causal structure} within a \textbf{global non-causal context} is essential for stable and efficient training.

\subsection{Sudoku: combinatorial constraint satisfaction}
\label{app:sudoku_details_rigorous}

\begin{figure}[ht]
    \centering
    \includegraphics[width=0.48\textwidth]{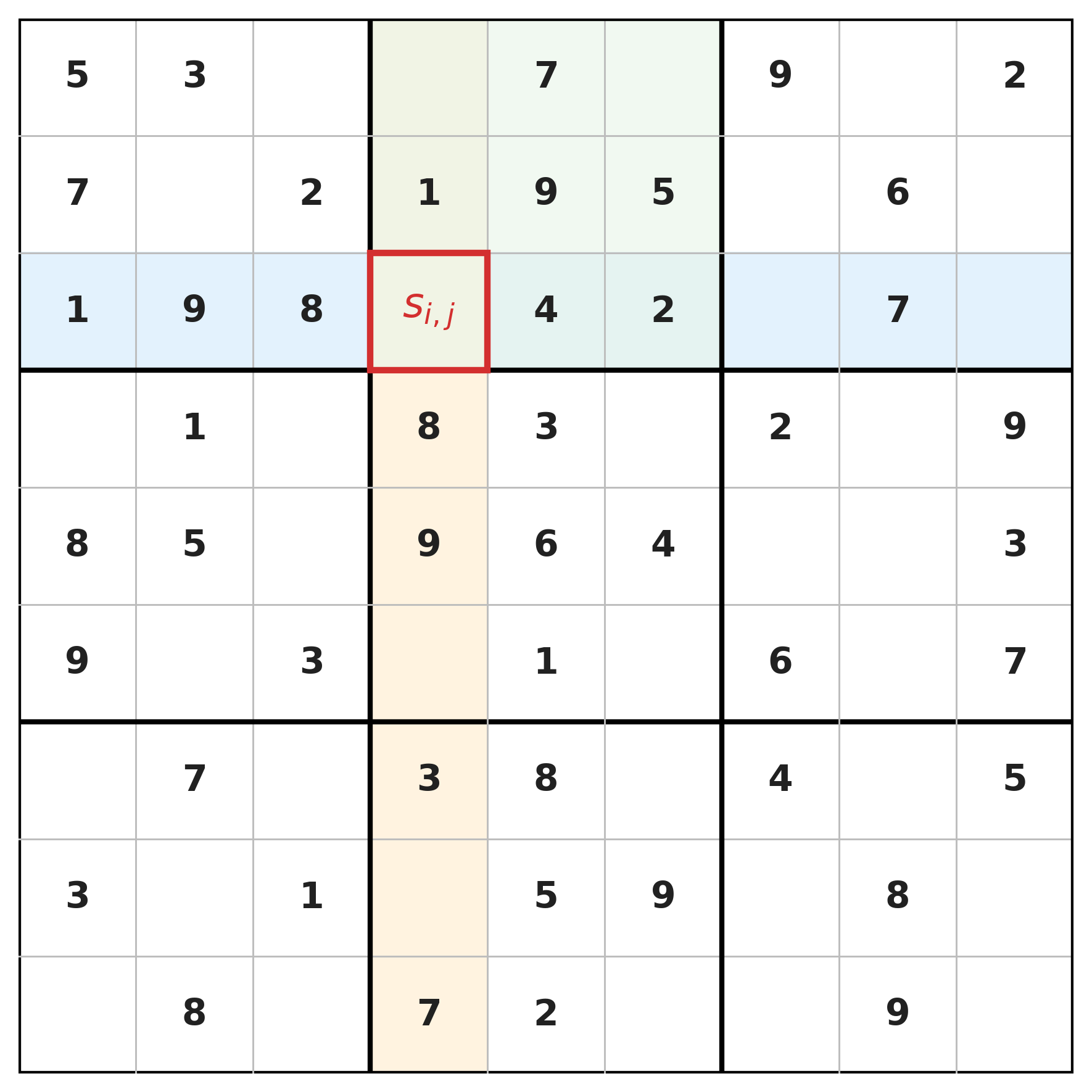} 
    \caption{Sudoku constraint domains. Each cell $x_{i,j}$ is governed by row, column, and block mutual exclusion rules, testing the model's ability to maintain global logical consistency.}
    \label{fig:sudoku_task}
\end{figure}

\paragraph{Task Formulation.} 
Sudoku serves as a probe for global constraint satisfaction. A puzzle is defined on a $9 \times 9$ grid where each cell must contain a digit $x_{i,j} \in \{1, \dots, 9\}$. The solution $\vx_{\text{res}}$ must simultaneously satisfy three algebraic mutual exclusion rules: each row, column, and $3 \times 3$ block must contain each digit exactly once. This creates a dense dependency network where each cell is coupled with 20 others (Fig.~\ref{fig:sudoku_task}).

\paragraph{Coordinate Awareness Implementation.} 
To restore the 2D spatial topology lost during 1D serialization, we implement a Tri-partite Coordinate Embedding mechanism. For a cell at flattened grid index $i \in \{0, \dots, 80\}$, its row ($r_i$), column ($c_i$), and $3 \times 3$ block ($b_i$) indices are derived as:
\begin{equation}
    r_i = \lfloor i / 9 \rfloor, \quad c_i = i \pmod{9}, \quad b_i = \lfloor r_i / 3 \rfloor \times 3 + \lfloor c_i / 3 \rfloor,
\end{equation}
where $r_i, c_i, b_i \in \{0, \dots, 8\}$. The model's embedding space $n_{embd}$ is partitioned into three segments to capture each constraint domain. The final coordinate representation $\mathbf{E}_{coord}(i)$ is the concatenation of learned embeddings:
\begin{equation}
    \mathbf{E}_{coord}(i) = [\text{Emb}_{row}(r_i) \parallel \text{Emb}_{col}(c_i) \parallel \text{Emb}_{blk}(b_i)].
\end{equation}

These coordinate embeddings are integrated into the Transformer backbone via an additive mechanism: $\mathbf{h}_j = \mathbf{h}_j + \mathbf{E}_{coord}(i)$, where $i$ is the grid position. This injection is applied to both the puzzle context ($\vx_{\text{prompt}}$) and the target response ($\vx_{\text{res}}$).

\paragraph{Evaluation Metrics.} 
\begin{itemize}[leftmargin=*]
    \item \textbf{Sudoku Accuracy}: Ratio of puzzles where the \textit{entire} $9 \times 9$ generated grid matches the ground truth.
    \item \textbf{Cell Accuracy}: Per-token precision calculated exclusively over the empty cells in the prompt $\vx_{\text{prompt}}$.
\end{itemize}

\begin{table}[ht]
\centering
\caption{Architectural and Diffusion Heuristics for Sudoku.}
\label{tab:sudoku_key_params}
\begin{tabular}{lll}
\toprule
\textbf{Parameter} & \textbf{Setting} & \textbf{Rational/Source} \\ \midrule
$n_{embd} / n_{layer}$ & $384 / 6$ & Shallow-but-wide for parallel search \\
Positional Bias & Coordinate IDs & Restoring 2D topology from 1D sequence \\
Jigsaw Criterion & Min-Entropy & Prioritizing high-certainty blocks \\
$T_{inf}$ (Inference) & $20$ steps & Budget for iterative refinement \\
\bottomrule
\end{tabular}
\end{table}

\subsection{Star-Graph path-finding: lookahead planning}
\label{app:pathfinding_details_rigorous}

\paragraph{Task Formulation.} 
This task isolates lookahead capacity where logical dependency flows backwards (Goal $\to$ Start). We construct synthetic star-graphs $G_{d,l}$ where a central node $v_{start}$ has $d$ outgoing branches of length $l$. Critically, only one branch terminates at $v_{goal}$, while others lead to dead ends. Transitions from $v_{start}$ are insufficient for greedy success (random accuracy $1/d$), requiring the model to trace the path from $v_{goal}$ back to $v_{start}$.

\paragraph{Serialization and Anisotropy.} Adjacency lists are randomly shuffled to remove positional cues: $\mathcal{X} = [\text{Edges} / v_{start}, v_{goal} =] \parallel [\text{Path Nodes}]$. 

\paragraph{The Jigsaw Anomaly.} Jigsaw fails here because its intra-block causal mask blocks visibility of the $v_{goal}$ constraint during the generation of the critical first token of any selected block, breaking the backward logical chain.

\paragraph{Evaluation Metrics.} 
Performance is evaluated using Path Accuracy (Sequence Exact Match). A path is considered correct if and only if the entire sequence of intermediate nodes matches the unique ground truth path. Any single node error results in 0\% for that sample.

\begin{table}[ht]
\centering
\caption{\textbf{Pathfinding Difficulty Configurations.}}
\label{tab:pathfinding_configs}
\begin{tabular}{lccccc}
\toprule
\textbf{Difficulty} & \textbf{Degree ($d$)} & \textbf{Length ($l$)} & \textbf{Nodes ($N$)} & \textbf{Train Samples}  \\ \midrule
Easy & 3 & 5  & 100 & 800K  \\
Moderate & 3 & 10  & 100 & 800K  \\
Standard & 5 & 5  & 100 & 800K  \\
Hard & 5 & 10 & 100 & 800K \\ \bottomrule
\end{tabular}
\end{table}

\subsection{Training and compute infrastructure}
All paradigms are trained using BF16 mixed precision and the AdamW optimizer. 
\begin{itemize}[leftmargin=*]
    \item \textbf{Diffusion schedule:} MDM, SDAR, and Scatter utilize a linear masking schedule $U(0.1, 0.9)$ with $T_{inf}=10$. 
    \item \textbf{Compute Cost:} Diffusion variants typically require $2 \times$ more total FLOPs than AR models to reach structural convergence, as reflected in the task affinity $A(t)$ in Fig.~\ref{fig:affinity-hardness-AR-LLMs-vs-dLLMs}.
    \item \textbf{Inference:} For Jigsaw, the intra-block causal mask is only applied during the "Solve" phase, while the "Probe" phase utilizes global bidirectional visibility for accurate entropy estimation.
\end{itemize}

\section{Ablation Study}
\label{app:ablation_study}

\subsection{The dependency conflict in path-finding}
\label{subsec:graph_bs_ablation}

In the main text (Sec.~\ref{sect:exp:path-finding}), we hypothesized that Jigsaw's failure on the Star-Graph task stems from a Dependency Conflict: the macro-policy selects goal-adjacent blocks first, but the micro-level intra-block autoregression forces blind left-to-right guessing within those blocks.

To rigorously validate this hypothesis, we conduct a fine-grained ablation on the block size $S \in \{1, 2, 4\}$. If our hypothesis holds, reducing $S$ to 1 should eliminate the intra-block autoregressive step entirely, thereby resolving the conflict.

\begin{figure*}[ht]
    \centering
    \includegraphics[width=0.98\textwidth]{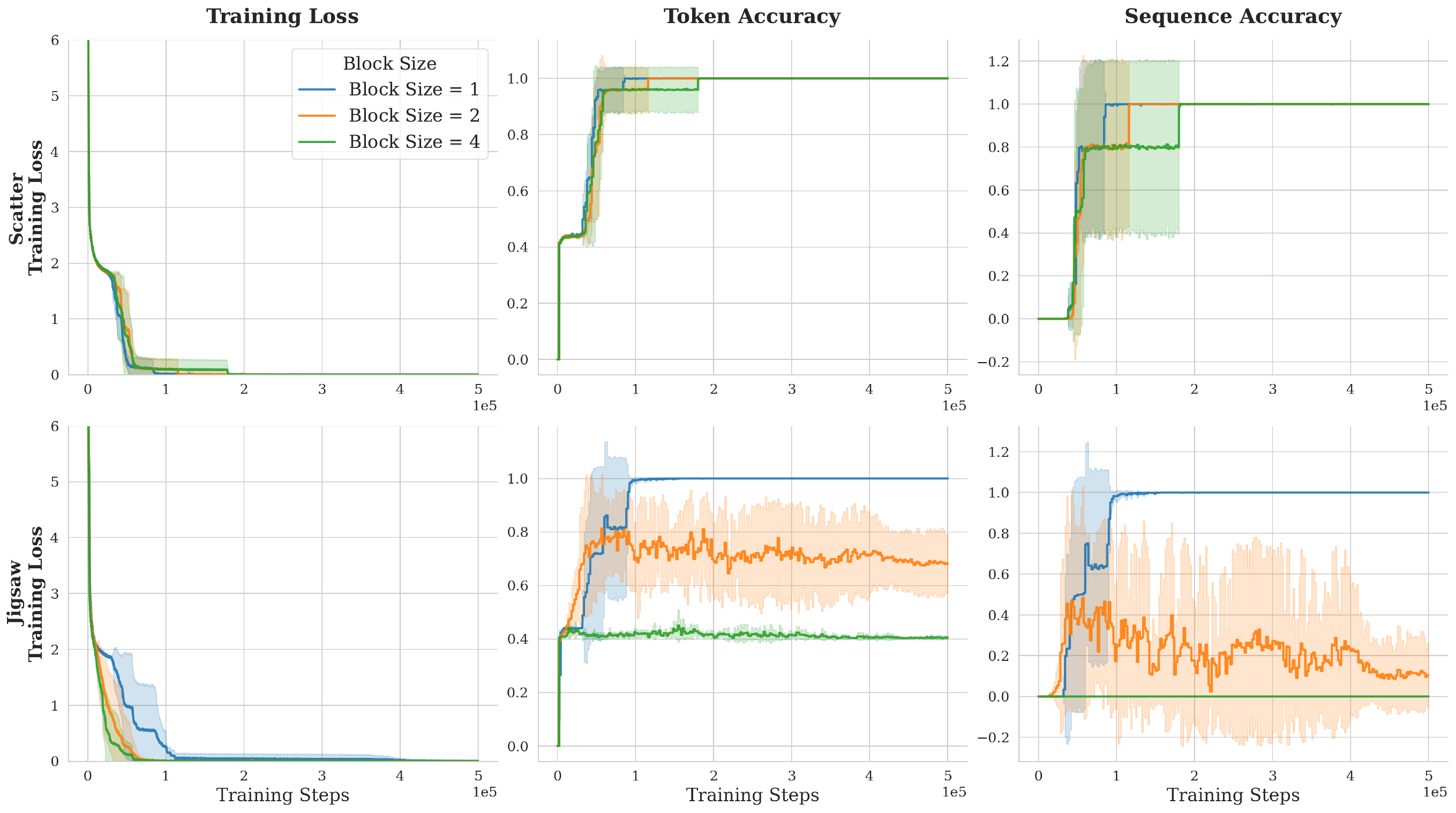}
    \caption{Ablation of Block Size ($S \in \{1, 2, 4\}$) on Path-Finding Dynamics.
    Top Row (Scatter): Demonstrates robustness to block size. Performance remains stable across all settings ($S=1$ marginally outperforms $S=4$), indicating that Scatter's synchronized offset mechanism effectively mitigates local dependency conflicts.
    Bottom Row (Jigsaw): Illustrates the critical sensitivity to locality granularity. At $S=4$ (green), the model fails completely. Reducing to $S=2$ (orange) yields partial success, while $S=1$ (blue) achieves near-perfect convergence, confirming that intra-block autoregression is the bottleneck.
    } 
    
    \label{fig:pathfinding_bs_ablation}
\end{figure*}

\paragraph{Scatter diffusion.}
The results presented in the top row of Figure~\ref{fig:pathfinding_bs_ablation} demonstrate that Scatter Diffusion retains performance stability across varying block sizes.
\begin{itemize}
    \item A minor performance trend of $S=1 \succ S=2 \succ S=4$ is observed, yet all configurations successfully converge to the solution.
    \item This resilience is attributed to Scatter's synchronized parallel generation mechanism. By evolving all blocks simultaneously rather than sequentially committing to them, Scatter allows global information to propagate backwards from the goal state via attention before intra-block tokens are fully determined. This circumvents the deadlock observed in Jigsaw.
\end{itemize}

\paragraph{Jigsaw diffusion} \mbox{} \\
The results presented in the bottom row of Figure~\ref{fig:pathfinding_bs_ablation} substantiate the dependency conflict hypothesis:
\begin{itemize}
    \item $S=4$ (Failure): A strong intra-block left-to-right constraint is imposed. Selection of goal-adjacent blocks results in the initial tokens lacking immediate predecessor context, leading to performance equivalent to random guessing (Sequence Acc $\approx 0.0$).
    \item $S=2$ (Transition): Reducing the block size mitigates the impact of uninformed prediction, enabling partial convergence, although training remains unstable.
    \item $S=1$ (Success): Critically, at $S=1$, the intra-block autoregressive process is effectively eliminated. The model relies solely on the entropy-based planner for token ordering. In this configuration, Jigsaw achieves near-perfect sequence accuracy ($\approx 1.0$), indicating the underlying planning algorithm is effective when not hindered by intra-block causal constraints.
\end{itemize}

\subsection{Effect of Non-Uniform Masking Schedules on Path-Finding}
\label{app:non_uniform_masking}

To determine whether the training instability of standard MDMs on the path-finding task (discussed in Section~\ref{sect:exp:path-finding}) can be resolved solely by modifying the training-time masking schedule, we conducted an ablation study using non-uniform masking distributions. 

For a target sequence of length $L$, we maintain the same total number of masked tokens $K$ as the uniform baseline, but sample the masked positions without replacement using position-dependent weights to inject directional bias:
\begin{itemize}[leftmargin=*]
    \item \textbf{Reverse-oriented bias:} Masks earlier tokens more frequently, keeping goal-side context visible. The sampling weight for position $i$ is set as $w_i^{\mathrm{rev}} \propto L - i + 1$.
    \item \textbf{Forward-oriented bias:} Masks later tokens more frequently. The sampling weight is $w_i^{\mathrm{fwd}} \propto i$.
\end{itemize}

\begin{figure}[h]
    \centering
    \includegraphics[width=0.9\textwidth]{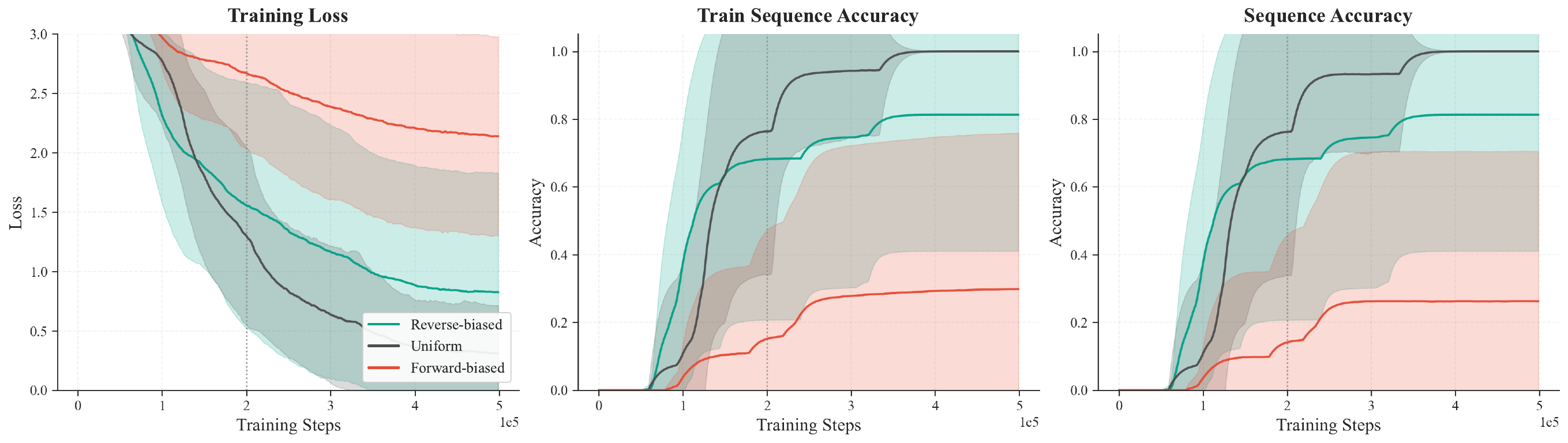}
    \caption{Training dynamics of path-finding under different masking schedules. While reverse-oriented bias accelerates early optimization, it leads to a lower final convergence ceiling compared to the uniform baseline.}
    \label{fig:non_uniform_masking}
\end{figure}

\textbf{Results and Discussion.} We observed that applying a reverse-oriented bias noticeably accelerates early-stage optimization compared to the uniform baseline, as it artificially increases the frequency of the necessary "right-to-left" learning signals. However, this directional bias merely alters the optimization trajectory without lifting the model's capability ceiling. By $200,000$ training steps, the uniform model catches up. Furthermore, imposing a strong directional bias causes over-specialization during training, which ultimately results in a \textit{lower} final convergence ceiling on the test set than the uniform baseline.

These results indicate that adjusting the noise schedule alone is insufficient to cure the structural mismatch. Explicit structural inductive biases---such as the synchronized parallelism introduced in Scatter---are necessary to robustly solve the task while retaining the high performance and flexibility of diffusion models.

\subsection{ICL Analysis: The capacity paradox in continuous induction}
\label{subsec:icl_scale_ablation}

In the functional induction task ($d=20$), the relationship between model scale and optimization efficiency is surprisingly non-monotonic.

\begin{figure}[ht]
\centering
\includegraphics[width=1.\textwidth]{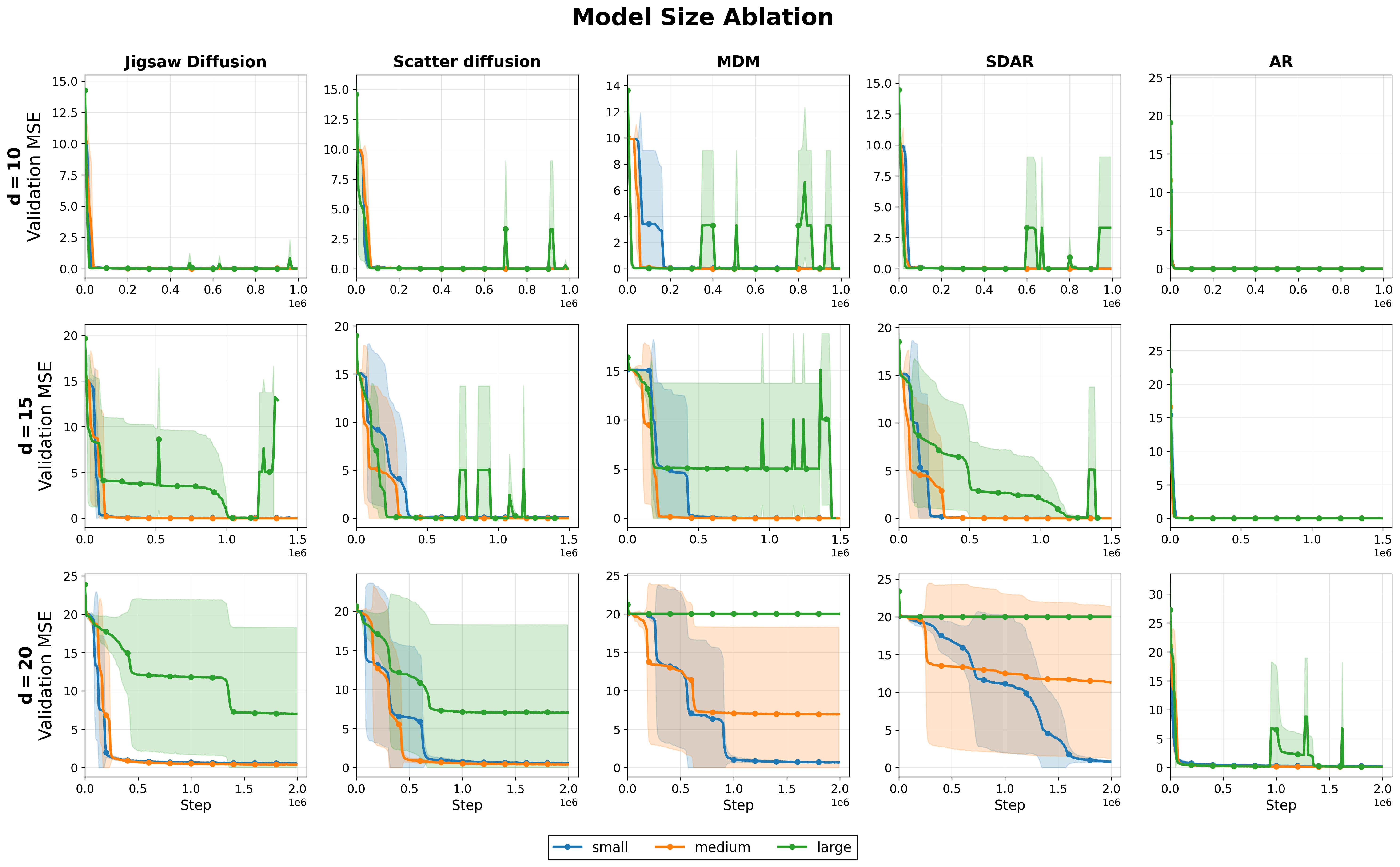}
\caption{\textbf{Validation MSE vs. Model Scale (ICL $d=20$).} Models follow LLaMA (AR) and LLaDA (MDM/blockwise) templates. Lower MSE indicates more precise mapping recovery. Small (blue) configurations identify the latent operator $\mW$ significantly earlier than Large (green) ones.}
\label{fig:icl_size_ablation}
\end{figure}

As shown in Fig. \ref{fig:icl_size_ablation}, Small and Standard configurations exhibit higher optimization stability. In high-dimensional linear regression, where the goal is to pinpoint a single linear mapping, the over-parameterized space of the Large model may introduce excessive gradient variance, delaying the "epiphany" jump. This suggests that for algebraic induction, architectural compactness can act as a regularizer.

\subsection{ICL Analysis: Locality-context balance: the block size ($S=4$) watershed}
\label{subsec:icl_bs_ablation}

Finally, we analyze the trade-off between local binding and global context in ICL.

\begin{figure}[ht]
\centering
\includegraphics[width=1.\textwidth]{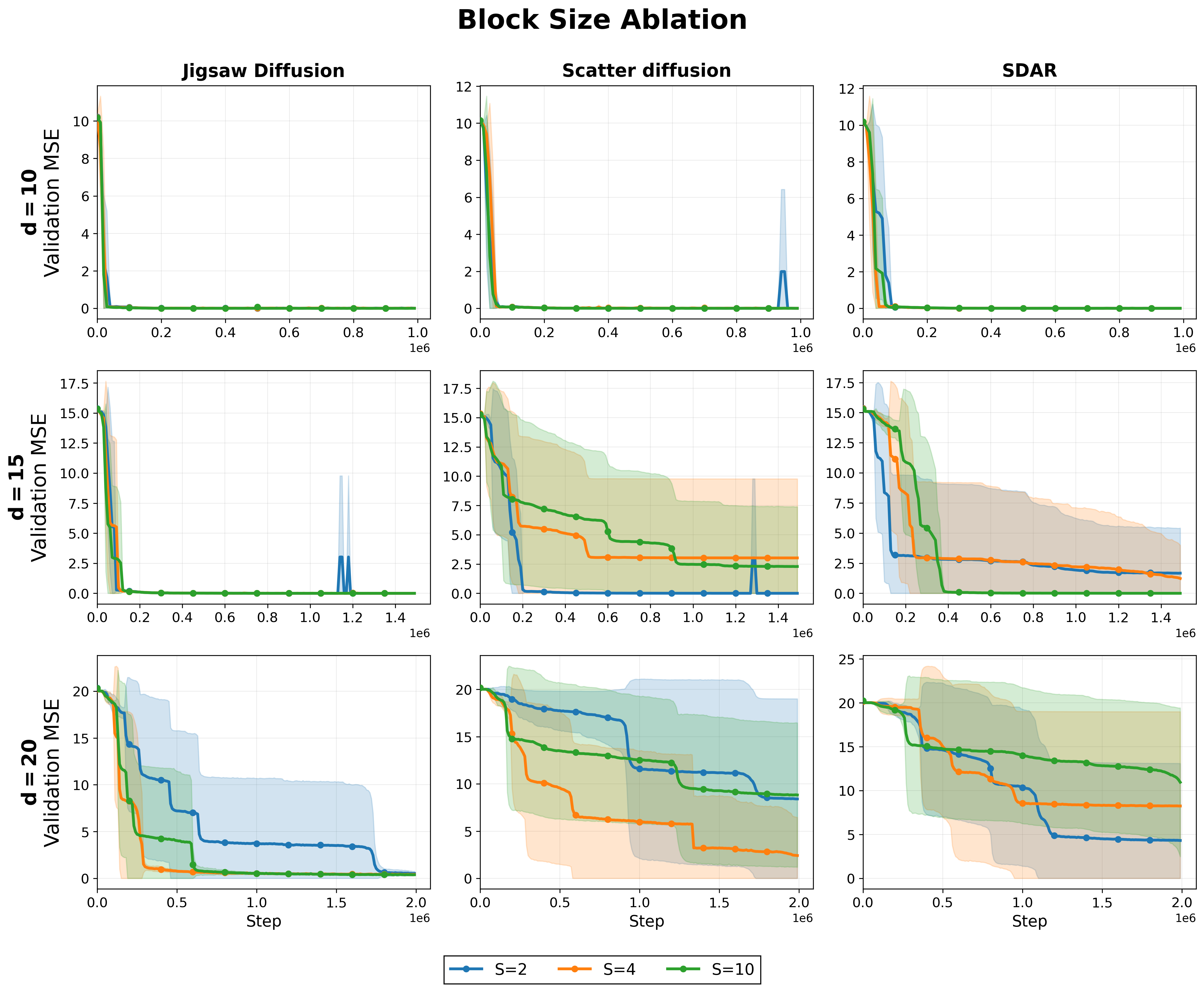}
\caption{\textbf{Locality Ablation for ICL MSE.} Across all dimensions, $S=4$ (orange) consistently serves as the "Goldilocks zone," providing the optimal balance between high-fidelity $(x, y)$ binding and global contextual reasoning.}
\label{fig:icl_bs_ablation}
\end{figure}

The results in Fig. \ref{fig:icl_bs_ablation} identify $S=4$ as the optimal granularity.
\vspace{-1em}
\begin{itemize}
    \item \textbf{Fragmentation ($S=2$):} Excessively small blocks fail to provide enough intra-block context for the model to "bind" features effectively, leading to slow MSE reduction.
    \item \textbf{Interference ($S=10$):} Excessively large blocks introduce destructive interference by mixing multiple, potentially conflicting prompt pairs into a single non-causal attention window, creating a performance floor that the model cannot overcome even with extended training.
\end{itemize}

\subsection{ICL Analysis: The Role of Data Diversity: Unique Samples vs. Multi-Epoch Training}
\label{subsec:icl_multi_epoch}

To determine whether the structural "epiphany" in ICL relies on total gradient steps or the diversity of demonstrations, we conduct a multi-epoch ablation study. We compare two training regimes with an identical compute budget (1M total steps):
\begin{itemize}
    \item \textbf{Unique Samples (E1):} Each training step utilizes a freshly sampled batch (1M unique task instances).
    \item \textbf{Limited Data (E10):} A fixed set of 100k task instances is sampled once and reused over 10 epochs.
\end{itemize}

\begin{figure*}[ht]
    \centering
    \includegraphics[width=1.\textwidth]{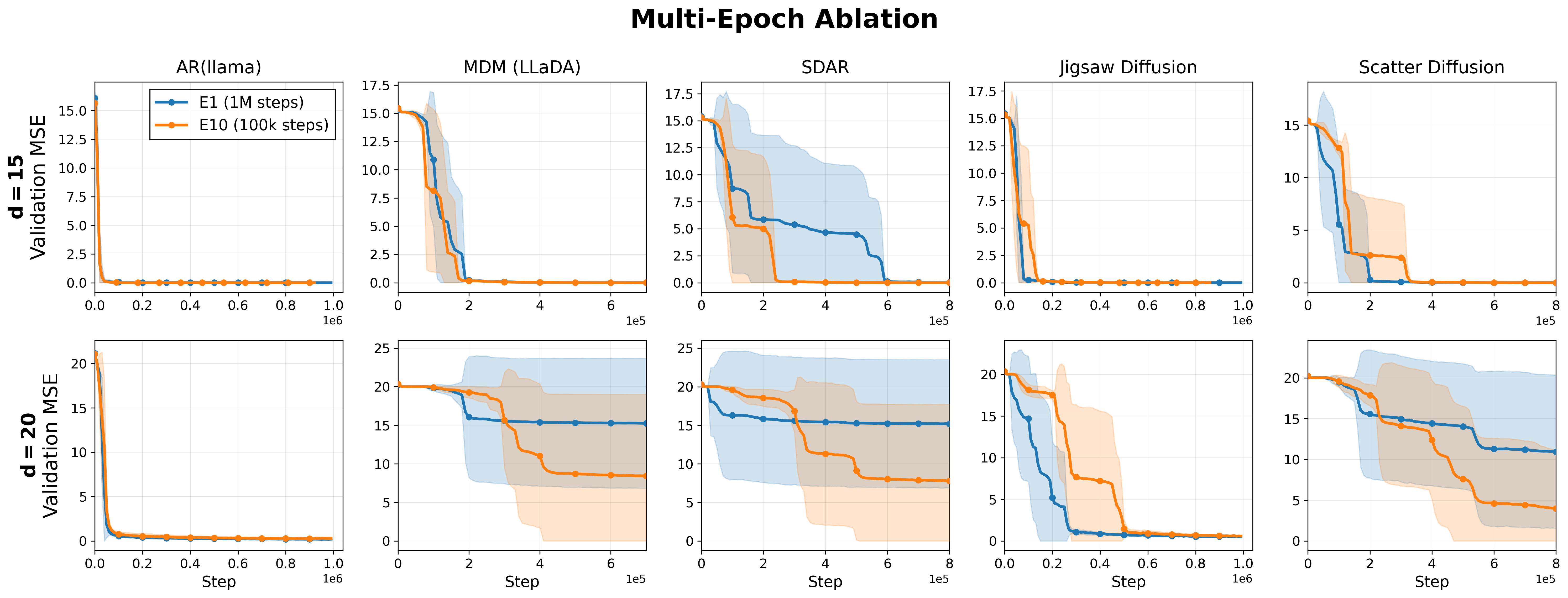} 
    \caption{\textbf{Multi-Epoch Ablation on ICL Validation MSE.} We compare 1M unique samples (\textbf{E1}, blue) against 100k samples repeated over 10 epochs (\textbf{E10}, orange). For the harder $d=20$ task, unique data diversity is critical for triggering the structural epiphany, whereas limited data causes delayed jumps or failed convergence in diffusion-based paradigms.}
    \label{fig:icl_multi_epoch}
\end{figure*}

\paragraph{Results Analysis.} As illustrated in Fig. \ref{fig:icl_multi_epoch}, the availability of unique samples significantly influences the optimization dynamics of diffusion paradigms, especially as dimensionality $d$ increases:
\begin{itemize}
    \item \textbf{Diversity-Driven Epiphany:} In the $d=20$ setting, models trained on unique samples (\textbf{E1}, blue) consistently exit the mean-prediction plateau earlier and more reliably. For Jigsaw Diffusion and Scatter Diffusion, the epiphany jump is markedly sharper and occurs at fewer total steps when the data is not recycled.
    \item \textbf{The Memorization Trap in High Dimensions:} In the E10 regime (orange), models often exhibit a "stagnant plateau." For MDM (LLaDA) at $d=20$, repeating the same 100k samples prevents the model from ever reaching structural recovery, causing it to stabilize at a high MSE. This suggests that with limited diversity, the model overfits to the specific noise patterns of the fixed Gaussian realizations rather than inducing the general linear mapping rule.
    \item \textbf{AR Robustness vs. Diffusion Sensitivity:} While the AR baseline is relatively robust to data recycling, diffusion-based paradigms (which rely on denoising global dependencies) require a continuous stream of novel task structures to resolve the gradient variance inherent in bidirectional induction.
\end{itemize}
In summary, data diversity acts as a crucial catalyst for algebraic induction; "seeing more tasks" is more effective than "seeing the same tasks more often" for overcoming the factorization curse in continuous spaces.

\subsection{Sudoku Analysis: Scaling laws and structural bottlenecks}
\label{subsec:sudoku_size_ablation}

We first examine whether the failure of autoregressive models on combinatorial tasks can be mitigated by increasing parameters. We evaluate three scales: Small (22M), Medium (67M), and Large (134M).

\begin{figure*}[ht]
    \centering
    \includegraphics[width=0.98\textwidth]{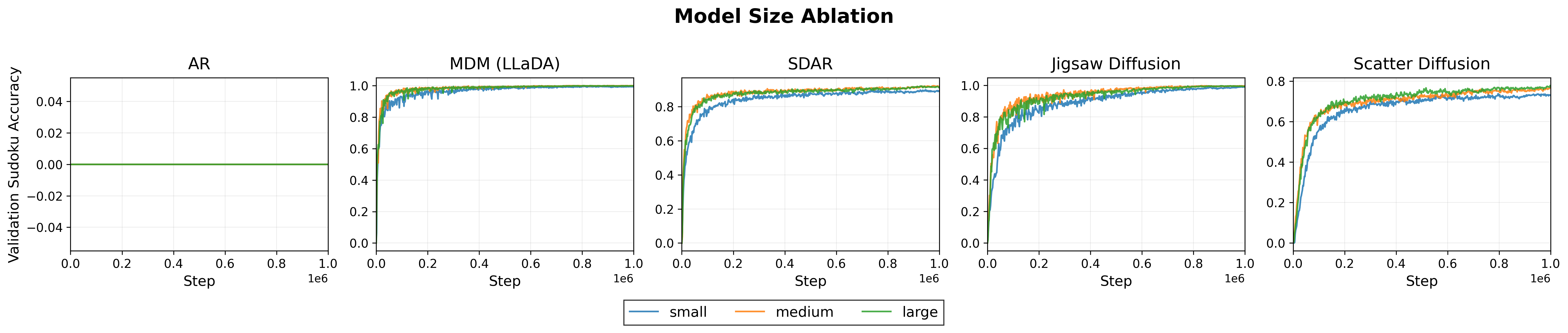}
    \caption{Validation Sudoku Accuracy across model scales. The AR baseline (gray) remains at 0\% accuracy regardless of scale, proving that the "Factorization Curse" is a structural deficiency. In contrast, diffusion paradigms (MDM, Jigsaw) exhibit clear scaling laws where larger models undergo an earlier "epiphany" phase and converge to higher solve rates.}
    \label{fig:sudoku_size_ablation}
\end{figure*}

\paragraph{Invariance of the Factorization Curse.} As shown in Fig. \ref{fig:sudoku_size_ablation}, the failure of the AR baseline is scale-invariant. Even at the Large scale, the model cannot overcome the compounding error inherent in the causal factorization of a non-causal solution space. Conversely, diffusion paradigms utilize increased capacity to better navigate the complex search space of the $9 \times 9$ grid, with Jigsaw showing the most efficient scaling behavior.

\subsection{Sudoku Analysis: Locality granularity ($S$)}
\label{subsec:sudoku_bs_ablation}

To understand how the size of the generation window affects constraint satisfaction, we test $S \in \{3, 9, 18\}$.

\begin{figure*}[ht]
    \centering
    \includegraphics[width=1.\textwidth]{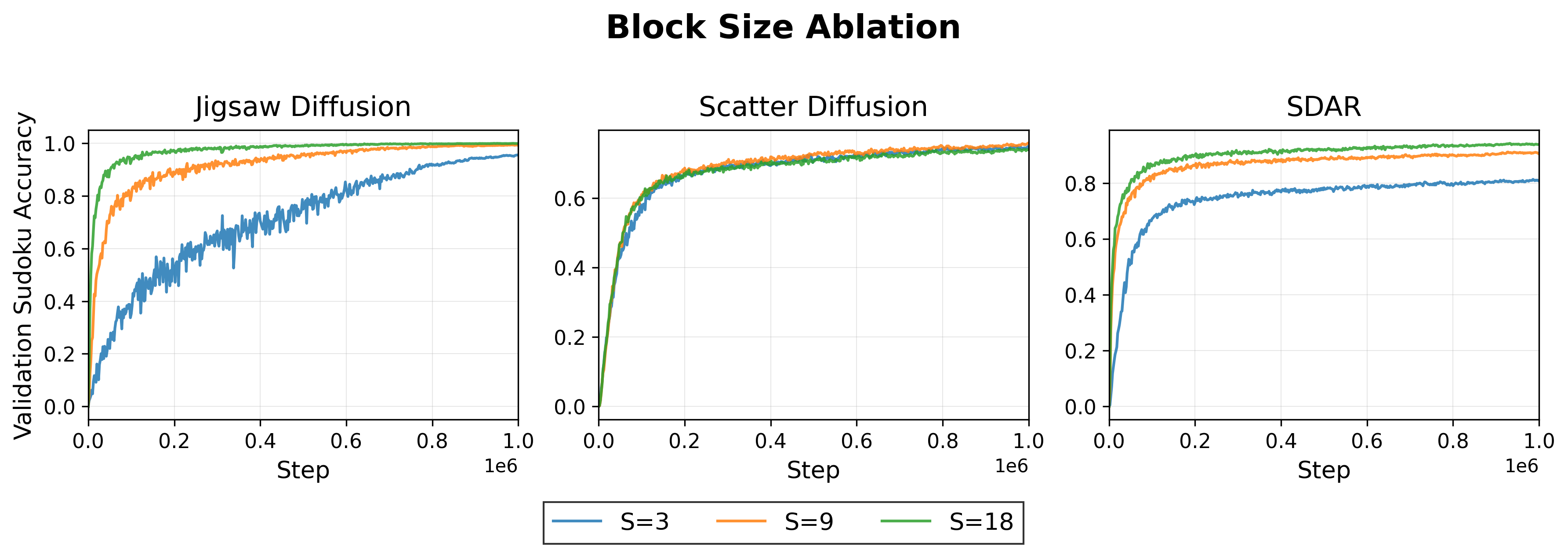}
    \caption{Impact of Block Size ($S$) on Sudoku Accuracy. Larger block sizes (e.g., $S=18$, green) provide superior stability for Jigsaw and SDAR by better encapsulating the row-wise mutual exclusion constraints of the grid.}
    \label{fig:sudoku_bs_ablation}
\end{figure*}

As illustrated in Fig. \ref{fig:sudoku_bs_ablation}, increasing $S$ from 3 to 18 leads to a significant reduction in training variance and faster convergence. For Jigsaw and SDAR, a larger block size permits the model to resolve entire constraint domains (e.g., two full rows) in a single synchronized step, preventing the "drift" that occurs when dependencies are resolved too incrementally.

\subsection{Sudoku Analysis: Impact of coordinate embeddings}
\label{subsec:sudoku_coord_ablation}

\begin{figure*}[ht]
    \centering
    \includegraphics[width=1.\textwidth]{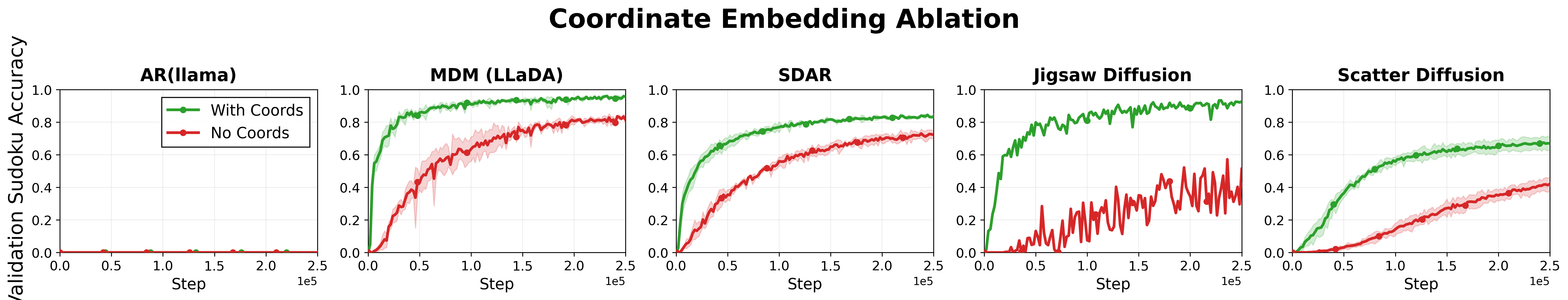}
    \caption{Coordinate Embedding Ablation on Sudoku. Performance comparison With Coords (green) and \textbf{No Coords} (red). Coordinate awareness serves as a crucial topological anchor, significantly accelerating convergence and raising the solve rate across all non-causal paradigms.}
    \label{fig:sudoku_coord_ablation}
\end{figure*}

To restore the 2D spatial topology lost during 1D serialization, we implement an additive Tri-partite Coordinate Embedding. For each cell, its row, column, and $3 \times 3$ block indices are mapped to learned embeddings and integrated into the Transformer: $h_j = h_j + [\text{Emb}_{row} \parallel \text{Emb}_{col} \parallel \text{Emb}_{blk}]$. We evaluate this topological bias across all paradigms in Fig. \ref{fig:sudoku_coord_ablation}.

\paragraph{Paradigm-Specific Dependency.} 
The ablation results (Fig. \ref{fig:sudoku_coord_ablation}) reveal a clear dependency hierarchy:
\begin{itemize}
    \item Planning Criticality (Jigsaw): Jigsaw Diffusion is most sensitive to spatial anchors. Without coordinates (red), the entropy-guided planner fails to identify constraint domains, leading to stagnant and volatile performance. With coordinates (green), it achieves a rapid "epiphany" and near-perfect accuracy.
    \item Convergence Acceleration (MDM, SDAR, Scatter): For other diffusion variants, coordinates provide a significant "head start" and a higher accuracy ceiling. By explicitly indexing the 20 relevant constraints for each cell, the attention mechanism more efficiently aggregates the discrete rules of the grid.
    \item Invariance of the Factorization Curse: The AR baseline remains at 0\% accuracy regardless of spatial information. This underscores that AR's failure is structural—rooted in its causal factorization bottleneck—rather than a lack of topological awareness.
\end{itemize}

\subsection{Sudoku Analysis: Inference step sensitivity}
\label{subsec:sudoku_nfe_ablation}

\begin{figure*}[ht]
    \centering
    \includegraphics[width=1.\textwidth]{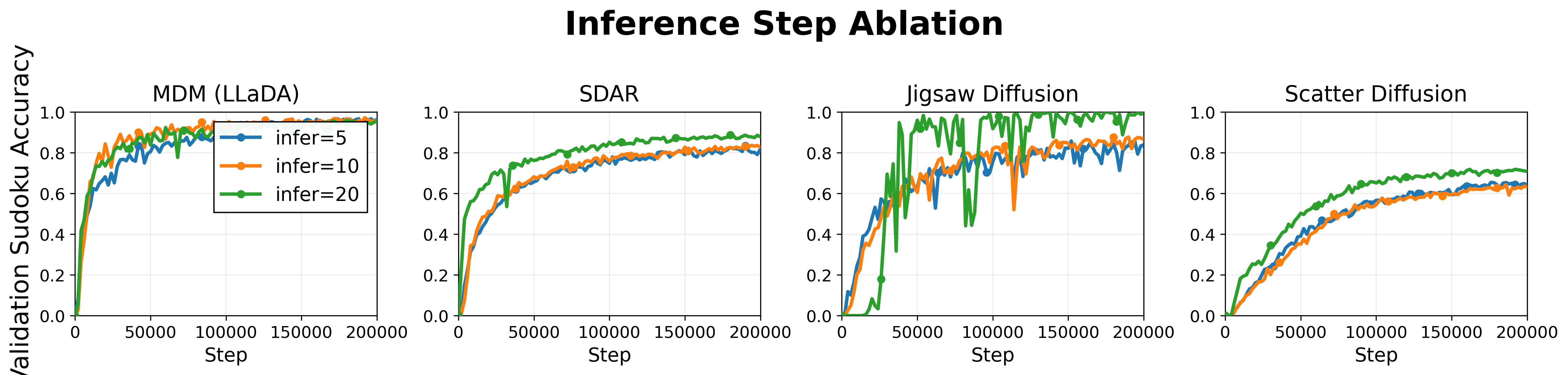}
    \caption{\textbf{Inference Step Ablation on Sudoku.} Curves represent $T=5$ (blue), $T=10$ (orange), and $T=20$ (green). MDM is highly efficient at low NFE, while blockwise paradigms (Jigsaw, SDAR, Scatter) scale positively with increased refinement steps.}
    \label{fig:sudoku_nfe_ablation}
\end{figure*}

We evaluate the impact of the number of function evaluations (NFE) by varying $T \in \{5, 10, 20\}$ during Sudoku inference (Fig. \ref{fig:sudoku_nfe_ablation}).

\vspace{2em}
\noindent \textbf{Results and Insights.} \\
The ablation reveals two distinct behaviors:
\begin{itemize}
    \item Global Efficiency (MDM): MDM (LLaDA) is exceptionally robust, reaching over 90\% accuracy even at $T=5$. This indicates that its global bidirectional attention can satisfy most grid constraints in very few refinement cycles.
    \item Blockwise Scaling (Jigsaw, SDAR, Scatter): These paradigms exhibit a clear performance hierarchy where $T=20 > T=10 > T=5$. For Jigsaw, $T=20$ is the critical threshold for reaching near-100\% accuracy. Similarly, Scatter and SDAR rely on higher NFE to provide the necessary iterative correction cycles to resolve inter-token conflicts within their synchronized prediction blocks.
\end{itemize}
In summary, while MDM excels in low-resource inference, blockwise paradigms leverage increased NFE to maximize precision in discrete constraint satisfaction.


\section{Analysis of Task Affinity and Compute Methodology}
\label{appendx-sect:fig-details}

To quantify the trainability and efficiency gaps across paradigms, we analyze the relative compute requirements as shown in Figure 1 of the main text. This section details the metrics and the underlying methodology for FLOPs estimation.

\subsection{Metrics: task affinity and hardness}

For a task $t$, let $C_{\mathrm{AR}}(t;\tau)$ and $C_{\mathrm{Diff}}(t;\tau)$ denote the cumulative training compute (in FLOPs) required by an AR model and a diffusion-based model to reach a target performance threshold $\tau$. 

\begin{itemize}
    \item Task Affinity ($A(t)$): $A(t) = \log_{10} \left( \frac{C_{\mathrm{AR}}(t;\tau)}{C_{\mathrm{Diff}}(t;\tau)} \right)$. A positive $A(t)$ indicates that the diffusion paradigm is more compute-efficient for that specific task.
    \item Compute Hardness ($H(t)$): $H(t) = \log_{10} \Big( \min\{C_{\mathrm{AR}}(t;\tau),\, C_{\mathrm{Diff}}(t;\tau)\} \Big)$. This represents the resources required by the most efficient paradigm to solve the task.
\end{itemize}

\subsection{Compute estimation methodology}
\label{app:compute_method}

We adopt a structured approach to estimate the FLOPs consumed during training and inference:

\paragraph{1. Profiling Strategy.} 
We measure the \texttt{train\_step\_flops} using the \texttt{fvcore} library [or specify your tool] at Step 10. This accounts for the specific architectural implementation (e.g., attention mechanisms and linear layers). We ensure that sequence lengths and batch sizes are identical across all compared paradigms for a specific task.

\paragraph{2. The Forward-Backward Decomposition.}
Total training compute is modeled based on the standard Forward-Backward relationship. Assuming the backward pass consumes approximately twice the compute of the forward pass, we derive the foundational unit forward compute ($\mathcal{F}$) from the total step compute ($\mathcal{C}_{step}$) as:
\begin{equation}
    \mathcal{F} = \frac{\mathcal{C}_{step}}{3}.
\end{equation}

\paragraph{3. Inference Complexity (Relative Analysis).}
While $A(t)$ in Figure 1 focuses on training efficiency, we provide the following model for inference to highlight the total deployment cost:
\begin{itemize}
    \item AR Models: $\mathcal{C}_{\text{inf, AR}} \approx \mathcal{F} \times n_{\text{respond}}$. (Assuming KV-caching is enabled).
    \item Diffusion Models: $\mathcal{C}_{\text{inf, Diff}} = \mathcal{F} \times \text{sampling\_steps}$. 
\end{itemize}

\subsection{Target performance thresholds ($\tau$)}
\label{app:thresholds}

The thresholds $\tau$ are task-specific and correspond to the points of "structural induction" or "successful solution." We define the following targets for Fig. 1:

\begin{table}[ht]
\centering
\caption{Target Thresholds ($\tau$) for Task Affinity Analysis.}
\label{tab:threshold_configs}
\begin{tabular}{lcc}
\toprule
\textbf{Task Category} & \textbf{Metric} & \textbf{Threshold $\tau$} \\ \midrule
ICL Linear Regression ($d=20$) & Validation MSE & 0.5 \\
ICL Linear Regression ($d=15$) &Validation MSE & 0.2 \\
Sudoku Solving & Accuracy & 95\% \\
Star-Graph Path-finding & Accuracy & 95\% \\ \bottomrule
\end{tabular}
\end{table}

\subsection{Interpolation and censoring}
\label{app:censoring}

The compute cost $C(t;\tau)$ is estimated via linear interpolation in the log-compute space. For models failing to reach the threshold $\tau$ within the maximum budget of $K_{\max}$ steps (e.g., $2 \times 10^6$), the cost is capped at the compute ceiling:
\begin{equation}
    C_{\max} = K_{\max} \cdot \mathcal{C}_{\text{step}}.
\end{equation}
In Figure 1, these censored points are marked with triangles, representing a conservative lower bound on the actual efficiency gap.

\end{document}